%% file: iclr2027_conference.tex
\documentclass{article} 
\usepackage{iclr2027_conference,times}

\input{math_commands.tex}

\usepackage{hyperref}
\usepackage{url}
\usepackage{graphicx}
\usepackage{booktabs}
\usepackage{multirow}
\usepackage{amsmath,amssymb,amsthm}
\usepackage{algorithm}
\usepackage{algorithmic}
\usepackage{tikz}
\usetikzlibrary{positioning,arrows.meta,shapes.geometric,fit}
\usepackage{xcolor}
\usepackage{subcaption}

\newcommand{\ours}{ZO-Curv}                 
\newcommand{\zoq}{q}                        

\theoremstyle{definition}
\newtheorem{theorem}{Theorem}
\newtheorem{proposition}{Proposition}
\newtheorem{lemma}{Lemma}
\newtheorem{assumption}{Assumption}

\newtheorem{remark}{Remark}
\newtheorem{corollary}{Corollary}
\makeatletter
\newcommand{\restated@name}{}
\newtheorem*{restated@thm}{\restated@name}
\newenvironment{restated}[2]{%
  \renewcommand{\restated@name}{#1~\ref*{#2}}\begin{restated@thm}}%
  {\end{restated@thm}}
\makeatother

\title{Accelerating LLM Pretraining with Zeroth-Order Curvature on Idle Serving Compute}

\iclrfinalcopy 

\author{
Mingchuan Ma$^{1,*}$,
Dongqi Cai$^{2,6,*,\dagger}$,
Daqian Yang$^{1}$,
Zhenyan Lu$^{3}$,
Chen Chen$^{4}$,
Teresa Lv$^{4}$\\
\bf
Yan Gao$^{5}$,
Shangguang Wang$^{3}$,
Mengwei Xu$^{2,3,6}$,
Nicholas D. Lane$^{5}$,
Yu Huang$^{1,\dagger}$ \\
$^1$ School of Software \& Microelectronics, Peking University \\
$^2$ Nanjing University \\
$^3$ Beijing University of Posts and Telecommunications \\
$^4$ Sony Europe Ltd. \\
$^5$ University of Cambridge \\
$^6$ MingTi Technology \\
}

\begin{document}

\maketitle

\begingroup
\renewcommand{\thefootnote}{}
\footnotetext{%
  \noindent\hspace*{-1.8em}%
  \begin{tabular}{@{}l@{\hspace{0.4em}}l@{}}
    $^*$ & These authors contributed equally to this work. \\
    $^\dagger$ & Corresponding author: cdq@nju.edu.cn, hy@pku.edu.cn
  \end{tabular}%
}
\endgroup

\input{sections/0_abstract}

\input{sections/1_introduction}
\input{sections/2_background}
\input{sections/3_method}
\input{sections/4_theory}
\input{sections/5_experiments}
\input{sections/6_related}
\input{sections/7_conclusion}

\input{sections/ai_use}
\input{sections/reproducibility}

\bibliography{iclr2027_conference}
\bibliographystyle{iclr2027_conference}

\appendix
\input{sections/8_appendix}

\end{document}

%% file: math_commands.tex
\usepackage{amsmath,amsfonts,bm}

\def\eqref#1{equation~\ref{#1}}

\def\1{\bm{1}}

\DeclareMathAlphabet{\mathsfit}{\encodingdefault}{\sfdefault}{m}{sl}
\SetMathAlphabet{\mathsfit}{bold}{\encodingdefault}{\sfdefault}{bx}{n}



%% file: sections/0_abstract.tex
\begin{abstract}
LLM pretraining is increasingly compute-intensive, while substantial inference serving compute remains idle during off-peak periods.
However, such serving hardware is difficult to use for conventional first-order training due to its memory, networking and other system requirements.
Zeroth-order (ZO) optimization offers a natural interface to this forward-only compute resource, but its gradient estimates break down in pretraining.
Unlike fine-tuning, which can confine updates to a small parameter subspace, pretraining must update every parameter, and the alignment of a ZO gradient estimate with the true gradient shrinks as the number of updated parameters grows.
We make a different observation: noisy ZO estimates can still provide useful curvature information, even when they are too inaccurate to serve as gradients.
Based on this insight, we propose ZO-Curv, which leverages idle forward-only devices to estimate curvature and uses it to adapt and refine the update used during first-order training, thereby accelerating the process.
We provide convergence guarantees that characterize the effect of finite-query curvature estimation.
Experiments on LLMs from 125M to 1.5B parameters show that ZO-Curv consistently accelerates pretraining.
At 770M parameters, ZO-Curv reaches AdamW's 20K-step validation loss in 3.9 hours instead of 7.5 hours, while achieving a lower final loss of 2.70 versus 2.84 under the same step budget.
These results show that idle model serving compute can accelerate LLM pretraining when zeroth-order queries are used to estimate curvature rather than gradients directly.
\end{abstract}

%% file: sections/1_introduction.tex
\section{Introduction}
\label{sec:intro}

LLM pretraining demands substantial compute, with model development requiring repeated runs over architectures, training schedules, data mixtures, and optimizer configurations~\citep{khandelwal2024100k,kaplan2020scaling,hoffmann2022chinchilla,zhang2022opt,yang2021tensor}.
Meanwhile, rapidly growing model serving compute, from inference clusters to local devices~\citep{cai2025federated}, represents a potentially valuable neglected resource for accelerating pretraining.

\paragraph{Serving compute is abundant but difficult to train on.}
Much of this serving capacity is underutilized.
As shown in Figure~\ref{fig:idle}(a), serving clusters are provisioned for peak demand, leaving substantial capacity idle during off-peak periods~\citep{patel2024splitwise,griggs2024melange,li2023lyra,liu2026valve}.
Across tens of thousands of production Kubernetes clusters, Cast AI reports only 5\% average GPU utilization from January 2025 to April 2026~\citep{castai2026optimization}.
However, this idle compute cannot readily join conventional first-order (FO) training.
Training must additionally store activations for backpropagation, gradients, and optimizer states, resulting in a $3.5$--$9\times$ memory overhead over inference (Figure~\ref{fig:bg-overhead}(a)).
It also requires optimized backward kernels, ultra high-speed networks and runtime support that serving platforms often lack~\citep{kwon2023efficient,zhong2024distserve}.

\paragraph{Zeroth-order optimization: promise and challenge.}
Zeroth-order (ZO) optimization needs only forward passes, estimating gradients from loss differences under random perturbations~\citep{spall1992spsa,nesterov2017random}.
This makes it a natural fit for forward-only hardware, and it has recently proven practical for \emph{fine-tuning} LLMs at inference-level memory cost~\citep{malladi2023mezo,zhang2024revisiting}.
Pretraining is harder: the variance of random-direction ZO estimators grows with the parameter dimension $d$~\citep{nesterov2017random,duchi2015optimal,shamir2017optimal}.
At the hundreds of millions to billions of parameters of modern LLMs, this leaves low-query estimates extremely noisy.
MeZO-style pretraining of GPT-2 355M confirms the effect, stalling near a validation loss of $8.5$ while AdamW drops below $3.0$ after the same number of training steps.
Mixing ZO estimates into the FO gradient inherits the same noise and underperforms AdamW (Figure~\ref{fig:idle}(d)).

\paragraph{Our approach: \ours{}.}
\ours{} uses ZO to estimate \emph{curvature} rather than gradients.
Unlike a gradient, curvature need not identify a precise descent direction, which noisy ZO queries cannot provide at pretraining scale; it only sets a coordinate-wise scale for the update, and even a coarse diagonal estimate can tell how aggressively each coordinate should move.
\ours{} therefore keeps exact FO gradient computation on the training node and outsources only curvature estimation to idle serving devices, which evaluate perturbed losses with forward passes.
The training node aggregates these estimates into a diagonal preconditioner for its FO update.

\input{floats/fig_bg_idle}

Our contributions are as follows.
\begin{itemize}
    \item We show that ZO estimates, while too noisy to provide reliable descent directions for LLM pretraining, remain effective for estimating curvature to precondition FO gradients.

    \item We propose \ours{}, a hybrid optimization framework that keeps exact FO gradient computation on the training node while offloading forward-only curvature estimation to idle serving devices.
    We further establish convergence guarantees for \ours{}.

\item We demonstrate consistent gains across models from 125M to 1.5B parameters.
\ours{} reaches AdamW's target loss up to $2.2\times$ faster, with larger models benefiting more from additional serving compute.
On the downstream task LAMBADA, \ours{} improves accuracy by $26.9\%$ over AdamW.

\end{itemize}

%% file: floats/fig_bg_idle.tex
\begin{figure}[t]
\centering
\includegraphics[width=\textwidth]{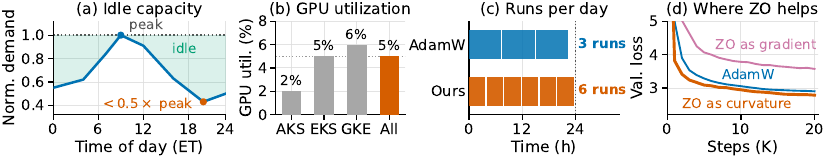}
\caption{
Landscape of idle serving devices and the pathway of harvesting them.
(a) Diurnal inference demand leaves peak-provisioned capacity idle.
(b) Production clusters exhibit low average GPU
utilization.
(c) Harvesting idle devices doubles the pretraining runs a cluster
completes per day.
(d) Comparison of ZO curvature and ZO gradients on pretraining.}
\label{fig:idle}
\end{figure}

%% file: sections/2_background.tex
\section{Background and Motivation}
\label{sec:background}

\subsection{Preliminaries}
\label{sec:prelim}

\paragraph{LLM pretraining.}
Pretraining minimizes the autoregressive negative log-likelihood
$F(\theta)=\mathbb{E}_{x\sim\mathcal{D}}\bigl[-\sum_{i}\log p_\theta(x_i\mid x_{<i})\bigr]$
over a large corpus $\mathcal{D}$, where $\theta\in\mathbb{R}^d$ are the transformer parameters.
The de facto optimizer is AdamW~\citep{loshchilov2019adamw}: every step runs a forward and a backward pass and maintains two moment buffers alongside parameters and gradients.

\paragraph{Second-order preconditioning.}
Sophia~\citep{liu2024sophia} accelerates pretraining by dividing the gradient EMA (Exponential Moving Average) by an EMA of a diagonal Hessian estimate, followed by coordinate-wise clipping.
The curvature estimate needs to be neither exact nor fresh: it is refreshed only every ten steps, and the EMA, the clipping, and a positive floor on the denominator absorb estimation noise.
This coarse signal halves the number of steps to reach a given validation loss.
Its cost, however, is charged to the training node, since the Gauss--Newton--Bartlett (GNB) estimator computes a minibatch gradient with resampled labels, an additional forward and backward pass.
Sophia accounts for this as $5\%$ extra compute per step on average, or roughly $50\%$ extra FLOPs on each refresh step.
Measured on GPU, however, a refresh step costs about twice an AdamW step rather than the $1.5\times$ that its FLOPs suggest (Section~\ref{sec:bg-overhead}).

\paragraph{Zeroth-order optimization.}
Zeroth-order (ZO) methods optimize with forward evaluations only~\citep{spall1992spsa,nesterov2017random}.
The two-point estimator perturbs the parameters along a random direction $u$ and forms
\begin{equation}
    \hat g \;=\; \frac{F(\theta+\varepsilon u)-F(\theta-\varepsilon u)}{2\varepsilon}\,u,
    \label{eq:two-point}
\end{equation}
which is unbiased for the gradient of a smoothed loss.
Its variance, however, grows linearly with the dimension $d$ at a fixed number of evaluations~\citep{nesterov2017random,duchi2015optimal}.
Because each evaluation needs only inference-level memory, ZO has recently proven practical for \emph{fine-tuning} LLMs~\citep{malladi2023mezo,zhang2024revisiting}, which raises the question of whether it can support pretraining as well.

\subsection{ZO gradients lose directional signal at pretraining scale}
\label{sec:bg-purezo}

\input{floats/fig_bg_purezo}

The most direct way to use forward-only serving devices is to let ZO replace backpropagation outright.
Figure~\ref{fig:bg-purezo}(a) shows that it does not work.
We pretrain a 355M model in GPT-2 structure from scratch on OpenWebText with a MeZO-style two-point estimator, against AdamW under an identical setup.
AdamW drops below a validation loss of $3.0$ within the budget, whereas pure ZO stalls near $8.5$ after an initial descent.
The gap is not a constant-factor slowdown: beyond the first few thousand steps, the ZO run makes essentially no further progress.

The failure is directional and scales with the number of trainable parameters $d$.
Figure~\ref{fig:bg-purezo}(b) compares the cosine similarity between the two-point estimate~\eqref{eq:two-point} and the exact gradient in pretraining ($d\approx3.5\times10^{8}$) and in LoRA fine-tuning of the same model on SST-2~\citep{socher2013sst} ($d\approx7.9\times10^{5}$).
Because the estimate is parallel to its Gaussian direction $u$, this cosine has expected magnitude close to $\sqrt{2/(\pi d)}$ for any gradient, matching both curves: it reflects the dimension, not the loss landscape.
Fine-tuning can confine the search to a small subspace, but pretraining must update every parameter, so each ZO estimate carries almost no directional information.
Forward-only devices should therefore supply not the descent direction but information that tolerates far noisier measurements.

\subsection{Serving devices cannot train, and curvature is costly}
\label{sec:bg-overhead}

\input{floats/fig_bg_overhead}

Could idle serving devices instead join conventional training as ordinary data-parallel workers?
Figure~\ref{fig:bg-overhead}(a) compares peak per-device memory for forward-only execution against AdamW and SophiaG training across model sizes.
Forward-only execution fits a representative $24$\,GB serving card at every size, peaking at $8.5$\,GB for the 1.5B model.
Training crosses that limit already at 355M and reaches roughly eight times the forward-only footprint at 1.5B, because it additionally holds activations for backpropagation, gradients, and optimizer state.
A device that comfortably serves a model still cannot train it.

Second-order preconditioning strains the training node further, more than its FLOPs accounting indicates.
Section~\ref{sec:prelim} reports Sophia consumes roughly $50\%$ extra FLOPs for the GNB refresh.
While measured on GPU (Figure~\ref{fig:bg-overhead}(b)), a step that recomputes the GNB estimate takes $2.12\times$ the time of an AdamW step.
Meanwhile, forward-only ZO estimation costs only $66\%$, and it runs within serving-level memory.
The compute that curvature estimation needs is cheap and abundant, but it sits on the wrong side of the cluster.

Together, the two negative results motivate a single split.
The first-order gradient is indispensable (Section~\ref{sec:bg-purezo}) and demands full training capability, so it must stay on the training node.
Curvature is auxiliary: it only rescales coordinates, the update rule that consumes it is built to tolerate noise (EMA, clipping, positive floor), and is friendly to ZO estimation on the idle serving nodes.

%% file: floats/fig_bg_purezo.tex
\begin{figure}[t]
\centering
\includegraphics[width=\textwidth]{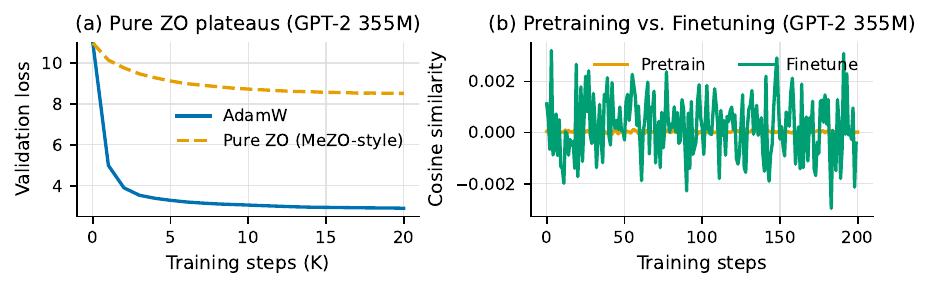}
\caption{ZO gradients fail at pretraining.
(a) Pure ZO stalls early while AdamW converges under the same
budget.
(b) Cosine similarity between the ZO gradient and the exact
gradient.}
\label{fig:bg-purezo}
\end{figure}

%% file: floats/fig_bg_overhead.tex
\begin{figure}[t]
\centering
\includegraphics[width=\textwidth]{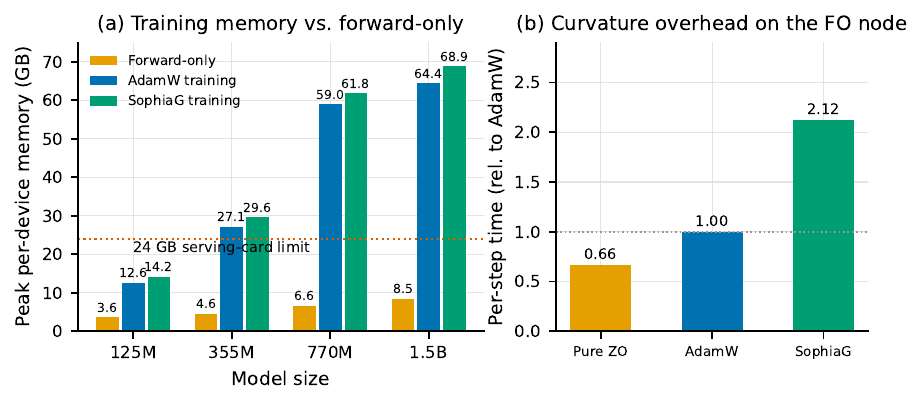}
\caption{Serving devices cannot join training, and curvature
estimation burdens the training node.
(a) Forward-only execution fits a $24$\,GB serving card at
every size; AdamW and SophiaG training exceed it from 355M onward.
(b) A SophiaG curvature-refresh step costs about twice an
AdamW step, while a forward-only computation is much cheaper to obtain.}
\label{fig:bg-overhead}
\end{figure}

%% file: sections/3_method.tex
\section{\ours{}: Curvature from Idle Forward-Only Compute}
\label{sec:method}


\subsection{Setting}
\label{sec:setting}

Figure~\ref{fig:overview} illustrates the overall \ours{} workflow.
We consider a service cloud with one strong training node and $M$
forward-only serving nodes. The training node supports backpropagation and
maintains optimizer states; serving nodes hold the model and evaluate losses
with inference-level memory. The nodes are physically interconnected within
the same cloud, and we assume that transferring model snapshots and query
results is not the dominant cost.

Let $F(\theta)=\mathbb{E}_{x\sim\mathcal D}[\ell(\theta;x)]$ be the
pretraining objective, with $\theta\in\mathbb R^d$.
The two pools run concurrently, offset by one step.
At step $t$, the training node broadcasts the snapshot $\theta_t$ to the
serving pool and computes the exact gradient $g_t=\nabla L_t(\theta_t)$
with true next-token labels; meanwhile, the pool evaluates its curvature
queries at $\theta_t$, whose results arrive in time for step $t+1$.
The update at step $t$ therefore uses the curvature estimated at
$\theta_{t-1}$: neither side ever waits for the other, and the curvature
EMA and the model are still refreshed at every training step.
The one-step offset has no measurable effect on training
(Appendix~\ref{app:pipeline-offset}).

\subsection{Forward-only diagonal curvature estimation}
\label{sec:estimator}

\input{floats/fig1_overview}

\paragraph{Estimation target.}
Our target is the Gauss--Newton--Bartlett (GNB) diagonal estimator used by
Sophia-G~\citep{liu2024sophia}.
Given a curvature minibatch $\mathcal B_t=\{x_b\}_{b=1}^B$, we sample
pseudo-labels $\hat y_b\sim p_{\theta_t}(\cdot\mid x_b)$ and define the
sampled-label loss
$\ell_t^{\mathrm{samp}}(\theta)=B^{-1}\sum_{b=1}^B
\ell_{\mathrm{ce}}(f_\theta(x_b),\hat y_b)$.
Writing $c_t=\nabla\ell_t^{\mathrm{samp}}(\theta_t)$, the reference estimate
is $h_t^{\mathrm G}=B\,c_t\odot c_t$.
Its expectation over the sampled labels recovers the minibatch
Gauss--Newton diagonal~\citep{martens2020new}. We estimate a damped version of this GNB
diagonal from forward passes only, avoiding the additional backward pass
needed to compute $c_t$.

\paragraph{Forward-only queries.}
At each step, the serving pool draws $q$ independent full-parameter
directions $u_{t,j}\sim\mathcal N(0,I_d)$ and evaluates two perturbed losses
per direction. All queries share the same snapshot, curvature minibatch,
and sampled labels, which are held fixed under perturbation. The returned
central differences determine the curvature estimate:
\begin{equation}
\begin{aligned}
    \Delta_{t,j}
    &=\frac{\ell_t^{\mathrm{samp}}(\theta_t+\varepsilon u_{t,j})
            -\ell_t^{\mathrm{samp}}(\theta_t-\varepsilon u_{t,j})}
           {2\varepsilon},\\
    \bar z_t&=\frac1q\sum_{j=1}^q\Delta_{t,j}u_{t,j},
    \qquad h_t^{\mathrm{ZO}}=B\,\bar z_t\odot\bar z_t.
\end{aligned}
\label{eq:main-zo-estimator}
\end{equation}
Averaging the directional estimates before squaring yields an unbiased estimate of the GNB diagonal with uniform damping $\|h_t^{\mathrm G}\|_1/q$ (Proposition~\ref{prop:main-curvature}).
The curvature minibatch and pseudo-labels are sampled on the serving nodes.
Thus, the downlink carries only the model snapshot and query indices, while the pool aggregates the directional terms internally.
A pool leader returns the averaged vector $\bar z_t$, requiring one gradient-sized transfer per refresh.
Overall, the estimator requires $2q$ perturbed forward evaluations across the pool.

\paragraph{Preconditioned update.}
The training node keeps two exponential moving averages, both initialized
at zero: the gradient momentum $m_t=\beta_1m_{t-1}+(1-\beta_1)g_t$ and the
curvature $v_t=\beta_2v_{t-1}+(1-\beta_2)h_{t-1}^{\mathrm{ZO}}$; the latter
uses the estimate of the previous step, which the pool returns one step
after receiving the snapshot (Section~\ref{sec:setting}).
Dividing the momentum by the floored curvature
$\hat d_t=\max\{\gamma v_t,\epsilon_0\}$ and clipping the result yields
\begin{equation}
    \theta_{t+1}=(1-\eta_t\lambda)\theta_t
    -\eta_t\operatorname{clip}(m_t/\hat d_t,\rho),
    \label{eq:zo-update}
\end{equation}
where $\gamma$ scales the curvature, $\lambda$ is the weight-decay
coefficient, and all operations act coordinate-wise.
The floor $\epsilon_0$ and the clipping threshold $\rho$ are what make this update
forgiving of curvature error: a positive denominator rescales a coordinate
without flipping its sign, and clipping caps the step wherever the
estimated curvature is too small.
Algorithms~\ref{alg:zocurv} and~\ref{alg:zocurv-serving} state both
procedures for a general staleness $s$, partitioning the $q$ query indices
into disjoint sets $\{\mathcal J_m\}_{m=1}^M$ across the pool.

\input{sections/alg}

\subsection{Convergence Analysis}
\label{sec:theory}

We study a deterministic core of~\eqref{eq:zo-update}: replace minibatch gradients
with $g_t=\nabla F(\theta_t)$ and omit momentum and weight decay; the
results hold for staleness $s$ (Remark~\ref{rem:staleness}).
Proofs and explicit constants are in Appendix~\ref{app:proofs}.

\begin{proposition}[Forward-only damped GNB curvature]
\label{prop:main-curvature}
Fix a step and condition on its curvature minibatch and sampled labels.
As $\varepsilon\to0$,
\begin{equation}
    \mathbb E[h_t^{\mathrm{ZO}}]=\tilde h_t
    :=\Bigl(1+\frac1q\Bigr)h_t^{\mathrm G}+\lambda_t\mathbf 1,
    \qquad
    \lambda_t=\frac{\|h_t^{\mathrm G}\|_1}{q},
    \label{eq:main-curvature-error}
\end{equation}
and $\mathbb E[(h_{t,i}^{\mathrm{ZO}})^2]\le C_v\tilde h_{t,i}^2$ for a
universal constant $C_v$.
\end{proposition}
The forward-only estimate is thus unbiased for a GNB diagonal with a
Levenberg--Marquardt-type damping $\lambda_t$ that decreases as $1/q$,
and its fluctuation is of the order of its target in every coordinate,
independently of $d$. Under Assumption~\ref{ass:hessian-lipschitz},
finite differences add a relative error of order $\varepsilon^2$.

Let $D_t=\operatorname{Diag}(d_t)$, $d_t=\max\{\gamma\tilde
v_t,\epsilon_0\}$, be the denominator built from the EMA $\tilde v_t$ of
$\tilde h_t$ at the same iterates, and let
$\widehat D_t=\operatorname{Diag}(\hat d_t)$ be its ZO counterpart.
Fresh directions at every refresh let the EMA average the fluctuations:
if each coordinate of $\tilde h_t$ varies by at most a factor $\kappa$
over the horizon, then with probability at least $1-\zeta$,
$(1-\delta)D_t\preceq\widehat D_t\preceq(1+\delta)D_t$ after
$\mathcal O(1/(1-\beta_2))$ warm-up steps, where
$\delta=\mathcal O\bigl(\kappa[\sqrt{(1-\beta_2)\Lambda}
+(1-\beta_2)\Lambda^2]+\sqrt{\zeta/T}+\varepsilon^2\bigr)$ and
$\Lambda=\log(qdT/\zeta)$ (Theorem~\ref{thm:query-dependent}); the
positive floor preserves this relative bound. We require $\delta<1$.

\begin{theorem}[Robust descent]
\label{thm:main-descent}
Suppose $\nabla^2F\preceq\alpha D_t$ along each update segment and the
above distortion bound holds. For $\eta\le(1-\delta)/\alpha$, the
deterministic update $\theta_{t+1}=\theta_t-
\eta\operatorname{clip}(\widehat D_t^{-1}g_t,\rho)$ satisfies
\begin{equation}
    F(\theta_{t+1})\le F(\theta_t)
    -\frac\eta2\sum_i\min\left\{
        \frac{g_{t,i}^2}{\hat d_{t,i}},\rho|g_{t,i}|\right\}.
    \label{eq:main-descent}
\end{equation}
If $F$ is bounded below, the minimum of this clipped stationarity measure
over $T$ steps is $\mathcal O(1/T)$.
\end{theorem}
A positive denominator preserves each gradient coordinate's sign, and
clipping bounds the update magnitude; smoothness and the step-size condition
ensure that the finite step decreases the loss.

\begin{theorem}[Local linear convergence]
\label{thm:main-linear}
Suppose clipping is inactive and the preconditioned PL condition
$g_t^\top D_t^{-1}g_t\ge2\mu_D(F(\theta_t)-F_\star)$ holds in the
local region. Under the conditions of Theorem~\ref{thm:main-descent},
choosing $\eta=(1-\delta)/\alpha$ gives
\begin{equation}
    F(\theta_{t+1})-F_\star
    \le\left(1-\frac{\mu_D}{\alpha}
                    \frac{1-\delta}{1+\delta}\right)
       (F(\theta_t)-F_\star).
    \label{eq:main-linear-rate}
\end{equation}
\end{theorem}
Curvature error enters only through the factor $(1-\delta)/(1+\delta)$,
which is governed by the EMA window and depends on $d$ only
logarithmically. The budget $q$ instead sets the damping of the
reference preconditioner: as $q$ grows, $\lambda_t$ decreases and $D_t$
approaches the undamped GNB preconditioner of Sophia-G.
The serving pool thus supplies coordinate-wise update scales while the
training node retains first-order gradient computation.

%% file: floats/fig1_overview.tex
\begin{figure}[t]
\centering
\includegraphics[width=0.95\linewidth]{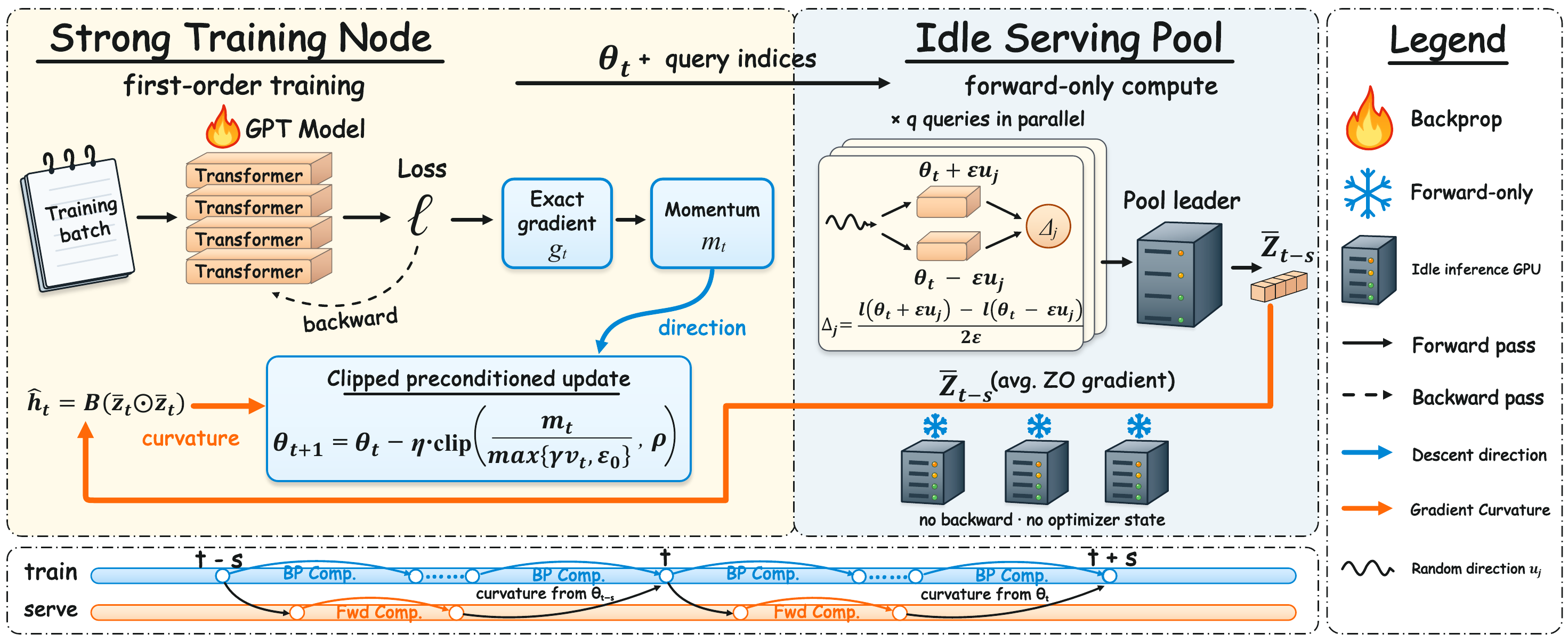}
\caption{\textbf{\ours{} overview.} The strong node runs ordinary
first-order pretraining; idle serving devices concurrently evaluate
$\zoq$ central-difference queries and return their averaged directional
estimate, from which the strong node assembles the diagonal curvature
that preconditions its update.}
\label{fig:overview}
\end{figure}

%% file: sections/alg.tex
\begin{figure}[t]
\captionsetup{hypcap=false}
\centering
\begin{minipage}[t]{0.55\linewidth}
\begin{algorithm}[H]
\caption{\ours{}: training node}
\phantomsection\label{alg:zocurv}
\small
\begin{algorithmic}[1]
\STATE \textbf{Input:} initial parameters $\theta_1$; learning rates
       $\{\eta_t\}_{t=1}^T$; query budget $q$; EMA coefficients
       $\beta_1,\beta_2$; curvature scale $\gamma$; clipping threshold
       $\rho$; denominator floor $\epsilon_0$; weight decay $\lambda$;
       staleness $s$
\STATE $m_0\gets0$; $v_0\gets0$;
       partition $\{1,\ldots,q\}$ into $\{\mathcal J_m\}_{m=1}^M$
\STATE Broadcast $(\theta_1,\{\mathcal J_m\})$;
       \ $t_{\mathrm{disp}}\gets1$
\FOR{$t=1,\ldots,T$}
    \STATE $g_t\gets\nabla L_t(\theta_t)$
    \STATE $m_t\gets\beta_1m_{t-1}+(1-\beta_1)g_t$
    \IF{$t-t_{\mathrm{disp}}=s$}
        \STATE $h\gets B\,\bar z_{t-s}^{\odot2}$
               \hfill{\footnotesize$\triangleright$ queries of
               $\theta_{t-s}$}
        \STATE $v_t\gets\beta_2v_{t-1}+(1-\beta_2)h$
        \STATE Broadcast $(\theta_t,\{\mathcal J_m\})$;
               \ $t_{\mathrm{disp}}\gets t$
               \hfill{\footnotesize$\triangleright$ non-blocking;
               pool runs Alg.~\ref{alg:zocurv-serving}}
    \ELSE
        \STATE $v_t\gets v_{t-1}$
    \ENDIF
    \STATE $\hat d_t\gets\max\{\gamma v_t,\epsilon_0\}$
    \STATE $\theta_{t+1}\gets(1-\eta_t\lambda)\theta_t
           -\eta_t\operatorname{clip}(m_t/\hat d_t,\rho)$
\ENDFOR
\end{algorithmic}
\end{algorithm}
\end{minipage}\hfill
\begin{minipage}[t]{0.43\linewidth}
\begin{algorithm}[H]
\caption{\ours{}: serving node $m$}
\phantomsection\label{alg:zocurv-serving}
\small
\begin{algorithmic}[1]
\STATE \textbf{Input:} snapshot $\theta_t$; query indices
       $\mathcal J_m$; finite-difference radius $\varepsilon$
\STATE Draw $\mathcal B$ and $\hat y\sim p_{\theta_t}$;
       form $\ell^{\mathrm{samp}}$
\FOR{$j\in\mathcal J_m$}
    \STATE $u_j\sim\mathcal N(0,I_d)$
    \STATE $L^{\pm}\gets\ell^{\mathrm{samp}}
           (\theta_t\pm\varepsilon u_j)$
    \STATE $\Delta_j\gets(L^{+}-L^{-})/2\varepsilon$
\ENDFOR
\STATE $z_m\gets\textstyle\sum_{j\in\mathcal J_m}\Delta_ju_j$
       \hfill{\footnotesize$\triangleright$ local partial sum}
\STATE Send $z_m$ to the pool leader
\STATE Leader: $\bar z_t\gets q^{-1}\sum_{m=1}^{M}z_m$
\STATE \textbf{Return} $\bar z_t$ to the training node
\end{algorithmic}
\end{algorithm}
\end{minipage}
\end{figure}

%% file: sections/4_theory.tex

%% file: sections/5_experiments.tex
\section{Experiments}
\label{sec:experiments}

We evaluate \ours{} to answer four research questions:
\textbf{RQ1:} does forward-only curvature estimation accelerate LLM pretraining end to end, in wall-clock time and across model sizes?
\textbf{RQ2:} how does the gain correlate with the query budget $\zoq$?
\textbf{RQ3:} can curvature queries substitute for first-order batch size, exchanging training-node compute for idle serving compute?
\textbf{RQ4:} do faster pretraining loss curves translate into downstream quality?

\subsection{Setup}
\label{sec:exp-setup}

\textbf{Models and data.}
GPT-2--style decoder-only transformers~\citep{radford2019gpt2} with 125M, 355M, and 770M parameters, plus a 1.5B GPT-NeoX configuration, trained from scratch on OpenWebText~\citep{gokaslan2019openwebtext} with context length 1024 and global batch size 512.
SophiaG and \ours{} train for 20K steps at 125M--770M and 50K steps at 1.5B; AdamW receives a doubled step budget (40K and 100K, respectively) so that all methods span a comparable wall-clock window.

\textbf{Baselines.}
(i)~AdamW~\citep{loshchilov2019adamw}, the FO standard.
(ii)~SophiaG~\citep{liu2024sophia}, which computes the GNB curvature by backpropagation on the training node, refreshed at every step so that both second-order methods consume equally fresh curvature.
(iii)~Pure ZO, a MeZO-style ZO-SGD~\citep{malladi2023mezo}.
(iv)~InstructedZO~\citep{cai2025federated}, which mixes ZO gradient estimates into the FO gradient and represents the gradient-level use of forward-only compute.
(v)~\ours{}, with $\zoq=512$ unless stated otherwise.
Further details are given in Appendix~\ref{app:details}.

\textbf{Hardware and timing.}
All experiments run on a single $8\times$H100 node.
The forward-only pool is emulated on the same hardware: at every step, the $2\zoq$ perturbed forward evaluations of Section~\ref{sec:estimator} are computed alongside the first-order step.
Wall-clock time for \ours{} counts the training-node critical path; the concurrent curvature evaluations are excluded, as they would occupy otherwise-idle serving devices.
Section \ref{sec:exp-main} additionally runs first-order training and the forward-only queries on disjoint GPU groups, and the speedup persists.

\subsection{RQ1: End-to-end acceleration}
\label{sec:exp-main}

\input{floats/fig2_main}
\input{floats/tab1_main}

Figure~\ref{fig:main} plots validation loss against wall-clock time and
grants AdamW twice the steps to span a comparable time window.
Table~\ref{tab:main} instead holds every method to the same 20K-step
budget at 770M and measures the time to reach AdamW's final loss.
\ours{} reaches this loss in roughly half the wall-clock time.
The speedup is consistent from 125M to 1.5B.

Table~\ref{tab:main} also shows that simply using forward-only compute
does not provide this benefit.
Pure ZO never reaches the target, while InstructedZO finishes above
AdamW.
Both use ZO estimates as descent directions.
In contrast, \ours{} uses forward-only estimates for curvature while
retaining the exact first-order gradient.
The benefit therefore comes from using forward-only compute for
curvature rather than for the descent direction.

The comparison with SophiaG further isolates \emph{where} curvature is
computed.
SophiaG and \ours{} make nearly identical progress per step
(Appendix~\ref{app:scaling}).
However, SophiaG computes its GNB curvature on the training node,
where a refresh costs $2.12\times$ an AdamW step
(Section~\ref{sec:bg-overhead}).
Table~\ref{tab:main} shows the resulting wall-clock overhead.
SophiaG reaches the target in only $10.0$K steps but takes $7.98$ hours,
more than the $7.53$ hours AdamW needs for its full run.
\ours{} reaches the same target in $10.5$K steps and takes only
$3.98$ hours.
Its curvature estimation runs concurrently on the idle serving pool,
leaving the training-node critical path close to that of AdamW.
Refreshing curvature every ten steps, as in the original Sophia,
amortizes the overhead only by using staler curvature.
Even then, SophiaG remains $6$--$12\%$ slower per step than \ours{}
(Appendix~\ref{app:sophia-k10}).
Offloading curvature removes this training-node overhead rather than
trading it against freshness.

Figure~\ref{fig:async-deploy} further validates the speedup under a
real concurrent deployment.
We partition one $8\times$H100 node into two disjoint four-GPU groups.
One group performs first-order training, while the other executes
forward-only curvature queries.
The returned curvature has a three-step staleness ($s=3$).
AdamW and SophiaG use only the four training GPUs, so the other four
GPUs represent otherwise-idle serving capacity.
Unlike the emulated setting above, the figure reports the measured
elapsed training time of the concurrent runs.
\ours{} completes 20K steps in about half the time required by SophiaG.
At 20K steps, \ours{} reaches a validation loss of $2.81$, close to
SophiaG's $2.78$, while requiring roughly half the elapsed training
time.
It also reaches this loss in less than half the time of AdamW.
Thus, the wall-clock acceleration persists when curvature estimation
and first-order training execute concurrently on disjoint GPU groups.

\subsection{RQ2: Effect of the query budget}
\label{sec:exp-query}

\input{floats/fig4_query}

In Section~\ref{sec:theory}, the damping of the curvature target decreases as $1/\zoq$, bringing the preconditioner closer to the GNB diagonal of SophiaG, and Figure~\ref{fig:query}(a,b) confirms the effect on the final loss.
The loss improves as $\zoq$ grows, monotonically until the curve flattens.
Where it flattens depends on scale: 125M is essentially saturated by $\zoq\approx32$ and 355M by $\zoq\approx256$, while 770M and 1.5B still improve at $\zoq=512$.
This ordering is consistent with the dimension dependence of the estimation error.
Figure~\ref{fig:query}(c,d) shows the corresponding trajectories at the two largest sizes.
Even $\zoq=4$ separates from AdamW within the first few thousand steps, and larger budgets lower the whole curve rather than only its endpoint.
Appendix~\ref{app:query-curves} repeats these panels for all four sizes.

The $\zoq$ queries are mutually independent, so the budget partitions freely across the pool and bounds how many idle devices can contribute in parallel.
Read as a provisioning rule, the sweep says that larger models absorb more idle capacity before the returns flatten.
Since a larger budget never costs accuracy, we use $\zoq=512$ in all other experiments.

\subsection{RQ3: Trading first-order batch for queries}
\label{sec:exp-batch}

\input{floats/fig56_row}

The experiments so far hold the first-order configuration fixed and add queries.
We now ask the converse: can queries pay for a \emph{smaller} first-order batch?
Figure~\ref{fig:heatmap} varies the training batch size jointly with $\zoq$ at 355M, with the step budget held fixed across cells and the $\zoq{=}0$ column giving plain AdamW.
Along the marked pairs, halving the batch and adding a modest query budget recovers the full-batch loss: a batch of 256 with $\zoq{=}32$ matches a batch of 512 at $3.03$.
Fewer queries are needed as the batch shrinks, eight when replacing a batch of 256 and a single one when replacing a batch of 128.
A first-order step at half the batch costs roughly half the training-node compute, so this substitution shortens every step while the added queries run on serving hardware.
\ours{} thus offers two ways to spend the same idle capacity: keep the batch and reach the target loss in less time (RQ1), or shrink the batch and cheapen every step at equal loss.
Both trade scarce training GPUs for abundant serving compute.

\subsection{RQ4: Downstream quality}
\label{sec:exp-downstream}

We evaluate the 770M checkpoints at the 20K-step budget of Table~\ref{tab:main} on HellaSwag~\citep{zellers2019hellaswag}, ARC-Easy and ARC-Challenge~\citep{clark2018think}, and LAMBADA~\citep{paperno2016lambada}.
We first evaluate the pretrained checkpoints with few-shot prompting and then apply LoRA finetuning to each task (Figure~\ref{fig:downstream}; protocol in Appendix~\ref{app:downstream}).
Before finetuning, \ours{} consistently outperforms AdamW across all four tasks and closely tracks SophiaG.
The largest gain appears on LAMBADA, where \ours{} improves accuracy from $30.8\%$ to $39.1\%$, a $26.9\%$ relative improvement over AdamW.

The advantage persists after finetuning.
\ours{} achieves an average score of $39.6$, compared with $35.2$ for AdamW and $39.8$ for SophiaG.
These results show that the lower pretraining loss of \ours{} translates into stronger downstream performance rather than remaining confined to the pretraining objective.
Meanwhile, its near-parity with SophiaG shows that curvature estimated on forward-only hardware provides comparable downstream quality to curvature computed by backpropagation.

%% file: floats/fig2_main.tex
\begin{figure}[t]
\centering
\includegraphics[width=\textwidth]{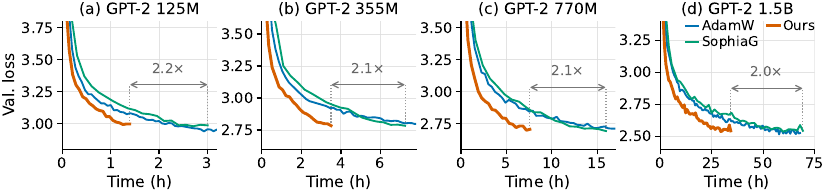}
\caption{Main result (RQ1): forward-only curvature turns step
efficiency into wall-clock speedup (OpenWebText; AdamW receives a
doubled step budget). \ours{} reaches AdamW's final loss in roughly
half the time at every size, whereas SophiaG spends its step advantage
on curvature computed on the training node.}
\label{fig:main}
\end{figure}

%% file: floats/tab1_main.tex
\begin{figure*}[t]
\centering

\begin{minipage}[t]{0.58\textwidth}
    \centering
    \vspace{-3cm} 

    \captionof{table}{Main comparison at GPT-2 770M with
    a shared 20K-step budget.
    Steps- and time-to-target measure how quickly each method
    reaches AdamW's final validation loss; ``--'' =
    never reached.}
    \label{tab:main}

    \resizebox{\linewidth}{!}{%
    \begin{tabular}{lccc}
    \toprule
    Method & Val.\ loss $\downarrow$ &
    \begin{tabular}{@{}c@{}}Steps to\\target (K) $\downarrow$\end{tabular} &
    \begin{tabular}{@{}c@{}}Time to\\target (h) $\downarrow$\end{tabular} \\
    \midrule
    AdamW                   & $2.84$ & $20.0$ & $7.53$ \\
    SophiaG                 & $2.69$ & $10.0$ & $7.98$ \\
    Pure ZO (MeZO-style)    & $8.01$ & --     & --     \\
    InstructedZO            & $3.37$ & --     & --     \\
    \textbf{\ours{} (ours)} & $2.70$ & $10.5$ & $\mathbf{3.98}$ \\
    \bottomrule
    \end{tabular}%
    }
\end{minipage}
\hfill
\begin{minipage}[t]{0.38\textwidth}
    \centering

    \includegraphics[width=0.8\linewidth]{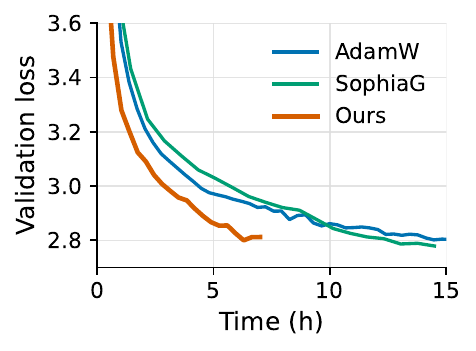}

    \captionof{figure}{Real deployment on disjoint GPU groups.
The x-axis reports the measured elapsed training time.}
    \label{fig:async-deploy}
\end{minipage}

\end{figure*}

%% file: floats/fig4_query.tex
\begin{figure}[t]
\centering
\includegraphics[width=\textwidth]{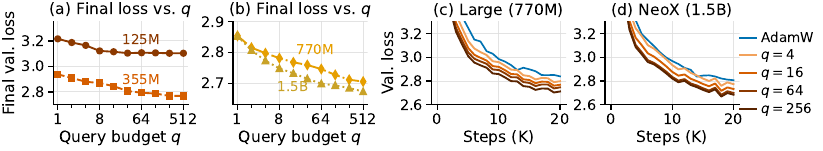}
\caption{Effect of the query budget $\zoq$ across model sizes.
(a)(b) Final validation loss for different query budgets; (c)(d) corresponding training trajectories.
}
\label{fig:query}
\end{figure}

%% file: floats/fig56_row.tex
\begin{figure}[t]
\begin{minipage}[t]{0.49\textwidth}
\centering
\includegraphics[width=\linewidth]{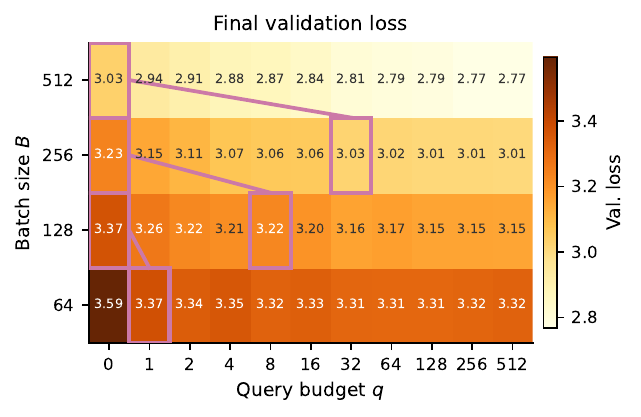}
\caption{Effect of the training batch size $B$ and query budget $\zoq$ on final validation loss for GPT-2 355M.
The $\zoq{=}0$ column corresponds to AdamW.}
\label{fig:heatmap}
\end{minipage}\hfill
\begin{minipage}[t]{0.49\textwidth}
\centering
\includegraphics[width=\linewidth]{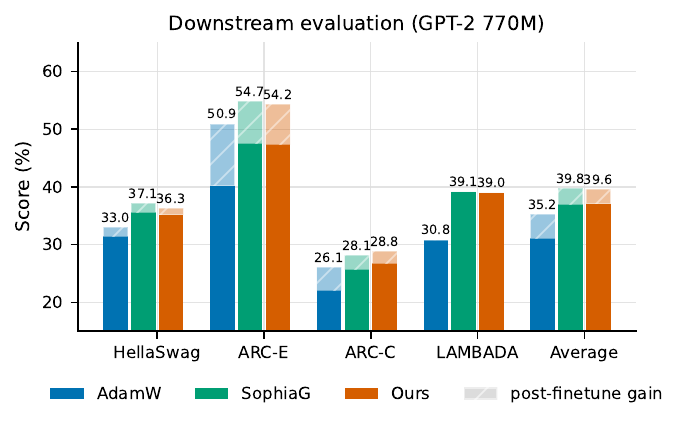}
\caption{Downstream performance of different models.
Solid bars: few-shot performance; hatched bars: LoRA finetuning gains.
}
\label{fig:downstream}
\end{minipage}
\end{figure}

%% file: sections/7_conclusion.tex
\section{Conclusion}
\label{sec:conclusion}
We show that idle serving compute can accelerate LLM pretraining by estimating curvature rather than gradients.
\ours{} keeps exact first-order gradients on the training node while using noisy zeroth-order estimates to precondition step sizes.
Across 125M--1.5B models, \ours{} matches SophiaG step for step and reaches AdamW's loss in roughly half the wall-clock time.
Future work will evaluate \ours{} on larger real-world serving clusters.

\newpage

%% file: sections/ai_use.tex
\subsection*{AI use statement}
In this work, we used generative AI tools to implement methods,
provide feedback on experimental design, interpret results, and
assist with translation. We did not use generative AI tools to
formulate the convergence results or to develop the core hypothesis
or conceptual framework of \ours{}. These contributions originated
with the authors. Generating synthetic datasets, cleaning or
reformatting datasets, and conducting qualitative data analysis
are not applicable to this work.
Additionally, we used generative AI tools to search for, identify,
and summarize related literature, to format references, and to
support brainstorming.
We have reviewed all AI-assisted work. AI-assisted implementations
were reviewed and validated through the training runs reported in
the paper. The authors verified the existence and bibliographic
details of every cited work and revised all AI-drafted text.
We take responsibility for the final content of this work,
including text, claims, or artifacts produced with the aid of
generative AI.

%% file: sections/reproducibility.tex
\subsubsection*{Reproducibility statement}
Section~\ref{sec:estimator} and Algorithms~\ref{alg:zocurv}
and~\ref{alg:zocurv-serving} specify \ours{} completely, including the
curvature estimator and the preconditioned update.
Section~\ref{sec:theory} summarizes our theoretical results, and
Appendix~\ref{app:proofs} states their assumptions and gives complete
proofs. Section~\ref{sec:exp-setup} describes the experimental protocol,
and Appendix~\ref{app:details} lists the model configurations, all
optimizer and estimator hyperparameters (Table~\ref{tab:hparams}), and
the profiling protocol. Our implementation builds on the public nanoGPT
code base used by Sophia~\citep{liu2024sophia}, and all data are public:
we pretrain on OpenWebText with the nanoGPT split and evaluate on
standard benchmarks following the protocol of
Appendix~\ref{app:downstream}. Appendix~\ref{app:async-deploy} describes
the deployment on disjoint GPU groups. We will release our code upon
acceptance.

%% file: sections/8_appendix.tex
\section{Related work}
\label{app:related}

\paragraph{Zeroth-order optimization.}
ZO methods build gradient surrogates from function evaluations alone,
from SPSA~\citep{spall1992spsa} to the smoothing analysis
of~\citet{nesterov2017random} and its stochastic nonconvex
extension~\citep{ghadimi2013stochastic}. Their rates degrade with the
dimension, and this dependence is optimal up to constants for two-point
estimators~\citep{duchi2015optimal,shamir2017optimal}.
MeZO~\citep{malladi2023mezo} made ZO practical for LLM fine-tuning by
regenerating perturbations from random seeds, and later work benchmarks
ZO fine-tuning~\citep{zhang2024revisiting} and reduces its variance with
adaptive momentum~\citep{jiang2024zo}, sparse
perturbations~\citep{liu2026sparse}, and low-rank
subspaces~\citep{chen2025enhancing,yu2025zeroth}.
DeepZero~\citep{chen2024deepzero} trains convolutional networks from
scratch with coordinate-wise estimates and pruning-guided sparsity, and
forward gradients~\citep{baydin2022gradients,ren2022scaling} replace
backpropagation with forward-mode directional derivatives, whose variance
likewise grows with the number of parameters.
InstructedZO~\citep{cai2025federated} mixes ZO estimates into
first-order gradients to pretrain on mobile devices. All of these use the
estimate as the descent direction; Section~\ref{sec:bg-purezo} shows
that this fails at LLM pretraining scale, and \ours{} uses ZO queries
only for curvature.

\paragraph{Zeroth-order curvature estimation.}
Second-order SPSA~\citep{spall2000adaptive} estimates the Hessian from
loss values alongside the gradient and uses it in a Newton-type update.
HiZOO~\citep{zhao2025second} estimates a diagonal Hessian with one extra
forward pass per step and uses it to precondition ZO fine-tuning of LLMs.
Both still take the descent direction from ZO estimates, in settings that
avoid backpropagation altogether. \ours{} instead pairs a forward-only
curvature estimate with the exact first-order gradient, targets the GNB
diagonal of Sophia-G for pretraining, and computes the estimate on
separate idle hardware.

\paragraph{Preconditioned optimizers for pretraining.}
Adaptive methods~\citep{kingma2015adam,loshchilov2019adamw} precondition
with gradient moments, Adafactor~\citep{shazeer2018adafactor} factors
these moments to save memory, and K-FAC~\citep{martens2015kfac} and
Shampoo~\citep{gupta2018shampoo} use structured curvature, with the
generalized Gauss--Newton matrix as the principled curvature for
likelihood losses~\citep{martens2020new}. More recent optimizers such as
SOAP~\citep{vyas2025soap} and Muon~\citep{liu2025muon} precondition LLM
pretraining with matrix-level structure. AdaHessian~\citep{yao2021adahessian}
and Sophia~\citep{liu2024sophia} instead use a diagonal Hessian estimate,
and Sophia shows that a clipped, EMA-smoothed estimate accelerates LLM
pretraining, but these estimators require extra backward passes on the
training node. The closest systems precedent
is Distributed Shampoo~\citep{anil2020scalable}, which computes the
inverse roots of its preconditioners asynchronously on otherwise
underused CPUs and refreshes them only every few hundred steps. Its
statistics, however, are still gathered from backpropagated gradients on
the accelerators, and only the matrix roots are offloaded. \ours{} moves
the curvature statistic itself to forward-only devices. It does not
propose a new preconditioner: it matches SophiaG step for step
(Appendix~\ref{app:scaling}).

\paragraph{Training on heterogeneous and idle compute.}
SWARM parallelism~\citep{ryabinin2023swarm} and
DiLoCo~\citep{douillard2023diloco} train across poorly connected devices,
Petals~\citep{borzunov2023petals} runs collaborative inference and
fine-tuning of large models, and Bamboo~\citep{thorpe2023bamboo} trains
on preemptible instances. Their training workers must all run
backpropagation. On the serving side, inference engines such as
vLLM~\citep{kwon2023efficient} and DistServe~\citep{zhong2024distserve}
are built around forward passes, and Splitwise~\citep{patel2024splitwise}
and M{\'e}lange~\citep{griggs2024melange} make inference clusters more
efficient. Lyra~\citep{li2023lyra} loans idle inference servers to
training jobs during low-traffic periods, Valve~\citep{liu2026valve}
colocates offline work with online serving, and Runway reclaims idle
inference GPUs for research jobs~\citep{dorsey2026borrowing}. These
systems hand idle serving capacity to other workloads, and loaned servers
must run full training. \ours{} instead lets forward-only capacity accelerate
the same pretraining run, without backpropagation on the serving devices.

\section{Experimental details}
\label{app:details}

\input{floats/tab_hparams}

\textbf{Models and data.} We build on the nanoGPT implementation used by
Sophia~\citep{liu2024sophia}: decoder-only transformers with GELU
activations, no bias terms, and no dropout, with the configurations of
Table~\ref{tab:hparams}. For the 355M and larger models we scale
attention by the inverse of the layer index, as Sophia does to keep
AdamW stable at these sizes. All models are trained from scratch on
OpenWebText~\citep{gokaslan2019openwebtext} with the GPT-2 tokenizer and
the train and validation split of nanoGPT (9B and 4.4M tokens), at
context length 1024 and global batch size 512, in bfloat16.

\textbf{Optimization.} All methods use a cosine learning-rate schedule
with 2K warmup steps, decaying to a minimum between $10^{-5}$ and
$6\times10^{-5}$, and gradient clipping at norm $1.0$. AdamW uses
$\beta=(0.9,0.95)$ and weight decay $0.1$. \ours{} and SophiaG use
$\beta=(0.965,0.99)$, weight decay $0.2$, clipping threshold $\rho=1$,
and the curvature scale $\gamma$ of Table~\ref{tab:hparams}, following
Sophia, and refreshes its curvature at every step. Step budgets are given in
Section~\ref{sec:exp-setup}.

\textbf{Curvature estimator.} Unless stated otherwise, \ours{} uses
$\zoq=512$ queries per refresh with finite-difference radius
$\varepsilon=10^{-4}$, and samples the pseudo-labels from the model's own
predictive distribution, as in the GNB estimator. Following the Sophia
implementation, the scale $B$ in $h^{\mathrm{ZO}}=B\,\bar z\odot\bar z$
counts the tokens of the curvature batch, and the denominator floor is
$\epsilon_0=10^{-15}$.

\textbf{Profiling.} The measurements of Section~\ref{sec:bg-overhead}
report peak reserved GPU memory at batch size 8, with forward-only
execution in fp16 and training in mixed precision, and per-step time
averaged over 100 steps after warmup.




\section{Step-axis comparison with SophiaG}
\label{app:scaling}

\input{floats/fig3_scaling}

Figure~\ref{fig:scaling} replots the runs of Figure~\ref{fig:main} on a
step axis. \ours{} and SophiaG are nearly indistinguishable at every
size, confirming that the forward-only estimator matches the
backpropagated GNB estimator in per-step optimization quality. The
contribution of \ours{} is not a stronger preconditioner but a
preconditioner obtained from hardware that could not otherwise
participate in training, which is why the two methods separate on the
wall-clock axis of Figure~\ref{fig:main}.

\section{SophiaG with a ten-step refresh interval}
\label{app:sophia-k10}

\input{floats/fig_sophia_k10}
\input{floats/fig_sophia_k10_loss}

Our main comparison refreshes the curvature of SophiaG at every step, so
that it consumes curvature as fresh as that of \ours{}
(Section~\ref{sec:exp-setup}). The original Sophia instead refreshes
every $k=10$ steps to amortize the cost of the GNB estimator.
Figure~\ref{fig:sophia-k10} reports the time each method needs for a
fixed budget of 20K steps under this setting, using the per-step
measurements behind Figure~\ref{fig:main}. A ten-step interval brings
SophiaG to $9$--$13\%$ above AdamW per step, while \ours{} stays within
$1$--$5\%$ of AdamW because its estimate is computed off the training
node. SophiaG therefore remains $6$--$12\%$ slower than \ours{} at every
size, and it obtains this reduction by refreshing its curvature ten
times less often, whereas \ours{} refreshes at every step.

The staler curvature also costs accuracy (Figure~\ref{fig:sophia-k10-loss}).
After 20K steps, SophiaG with a ten-step interval ends $0.03$--$0.04$
above its every-step counterpart in the main runs and $0.02$--$0.04$
above \ours{} at every size. Amortizing the estimator on the training
node thus leaves SophiaG both slower and less accurate than \ours{},
which keeps the curvature fresh at nearly the per-step cost of AdamW.

\section{Training trajectories across query budgets}
\label{app:query-curves}

\input{floats/fig7_query_curves}

Figure~\ref{fig:query-curves} extends the trajectory panels of
Figure~\ref{fig:query}(c,d) to all four model sizes. The qualitative
picture is uniform: every query budget separates from AdamW early, and
increasing $\zoq$ lowers the entire trajectory. The sizes differ only in
where the budgets stop mattering, consistent with the saturation points
of Figure~\ref{fig:query}(a,b).

\section{Effect of the one-step pipeline offset}
\label{app:pipeline-offset}

\input{floats/fig9_pipeline_offset}

The pipeline of Section~\ref{sec:setting} applies the curvature
estimated at $\theta_{t-1}$ to the update at step $t$.
Figure~\ref{fig:pipeline-offset} compares this deployed variant against
a synchronous control that waits for the curvature of the current
snapshot $\theta_t$. The two loss curves coincide throughout training:
consecutive snapshots differ by a single clipped update, and the
curvature EMA smooths over far longer horizons than one step, so the
offset costs nothing while removing the serving pool from the critical
path.

\section{Full downstream results}
\label{app:downstream}

\input{floats/tab2_downstream}

Table~\ref{tab:downstream} lists the per-task accuracies behind
Figure~\ref{fig:downstream}. We evaluate the 770M checkpoint of each
method after 20K steps, the shared budget of Table~\ref{tab:main}.

\textbf{Before finetuning.} We use few-shot prompting with 10 in-context
examples for HellaSwag, 25 for ARC-Easy and ARC-Challenge, and none for
LAMBADA, drawn from each training split with a fixed seed. For the three
multiple-choice tasks, the prediction is the choice with the highest
average log-likelihood per answer token, evaluated on the validation
split ($10{,}042$, $570$, and $299$ examples). LAMBADA is scored by
whether greedy decoding recovers the final word, on the $5{,}153$
test examples of the OpenAI version.

\textbf{After finetuning.} Starting from each checkpoint, we finetune
separately on the training split of every multiple-choice task with LoRA
(rank $8$, $\alpha=16$, dropout $0.05$) on the attention and MLP
projections, which trains $5.9$M parameters ($0.76\%$ of the model). We
use AdamW with learning rate $2\times10^{-5}$ and batch size $8$ for
$2{,}000$ steps, with the loss restricted to the answer tokens, and then
evaluate on the validation split. LAMBADA has no public
training split, so it is not finetuned.

\textbf{Results.} Before finetuning, \ours{} improves over AdamW on
every task, by $6.0$ points on average ($37.1$ versus $31.1$). It stays
within $0.4$ points of SophiaG on HellaSwag, ARC-Easy, and LAMBADA, and
is $1.0$ point higher on ARC-Challenge. Finetuning helps the AdamW
checkpoint most, by $4.1$
points on average against $2.8$ for SophiaG and $2.5$ for \ours{}, with
the largest gain on ARC-Easy ($10.7$ points). It nonetheless trails both
second-order checkpoints on every task after finetuning, while \ours{}
remains within $0.8$ points of SophiaG throughout ($39.6$ versus $39.8$
on average).

\section{Proofs}
\label{app:proofs}

This appendix provides the full theoretical analysis of ZO-Curv.
Our goal is to isolate the effect of replacing the backpropagation-based
Gauss--Newton--Bartlett (GNB) curvature estimator in Sophia-G
with a forward-only zeroth-order estimator.
The analysis proceeds in three stages.
First, we identify what the average-then-square ZO estimator estimates:
under a finite query budget $q$, it is an unbiased estimate of a damped
GNB diagonal whose damping decreases as $1/q$, and its fluctuation is of
the order of this target in every coordinate, independently of the
dimension $d$.
Second, we show that the curvature EMA averages these fluctuations into
a bounded multiplicative perturbation of the damped reference
preconditioner, and that such a perturbation does not destroy descent:
the update direction remains a descent direction and converges to
stationarity.
Third, we combine the two results into a convergence guarantee relative
to the damped-GNB-preconditioned reference update.

As in the theoretical analysis of Sophia \citep{liu2024sophia}, we analyze
a deterministic clipped preconditioned core in order to isolate the role
of curvature estimation. In particular, we omit gradient momentum and
decoupled weight decay from the convergence proof. The curvature EMA is
retained and updated at every iteration, matching the per-step estimator in Section~\ref{sec:method}.
We first analyze the synchronous core, in which the update at step $t$
uses the curvature queried at $\theta_t$; Remark~\ref{rem:staleness}
shows that the results carry over unchanged to the pipelined update with
any staleness $s$.

\subsection{Notation and reference curvature estimator}
\label{app:proofs:notation}

Let
\begin{equation}
    F(\theta)
\end{equation}
denote the training objective, with parameters
$\theta\in\mathbb{R}^{d}$.
At iteration $t$, let
\begin{equation}
    g_t := \nabla F(\theta_t)
\end{equation}
denote the first-order gradient used to update the model.

The curvature estimator is computed from a sampled-label mini-batch,
as in the GNB estimator of Sophia-G. Given a mini-batch
$\{x_b\}_{b=1}^{B}$, let
\begin{equation}
    \hat y_b
    \sim
    \operatorname{softmax}(f(\theta_t,x_b)),
    \qquad b=1,\ldots,B,
\end{equation}
and define the sampled-label loss
\begin{equation}
    \ell_t^{\mathrm{samp}}(\theta)
    :=
    \frac{1}{B}
    \sum_{b=1}^{B}
    \ell_{\mathrm{ce}}
    \bigl(f(\theta,x_b),\hat y_b\bigr).
\end{equation}
We denote its gradient at $\theta_t$ by
\begin{equation}
    c_t
    :=
    \nabla
    \ell_t^{\mathrm{samp}}(\theta_t).
    \label{eq:curvature-gradient}
\end{equation}
Conditioned on the sampled mini-batch and pseudo-labels, $c_t$ is fixed
throughout the $q$ ZO queries.

The backpropagation-based GNB estimator used as our reference is
\begin{equation}
    h_t^{\mathrm{G}}
    :=
    B\,
    c_t\odot c_t,
    \label{eq:raw-gnb}
\end{equation}
where $\odot$ denotes element-wise multiplication.
Taking expectation over the sampled labels recovers the diagonal of
the mini-batch Gauss--Newton matrix, as established by the Bartlett
identities in \citep{liu2024sophia}. Our analysis below compares the ZO
estimate with this reference conditioned on the same mini-batch and the
same sampled labels.

A finite query budget adds a uniform damping to this target
(Proposition~\ref{prop:main-curvature}). We therefore define the
damped GNB diagonal
\begin{equation}
    \tilde h_t
    :=
    \left(1+\frac{1}{q}\right)h_t^{\mathrm{G}}
    +
    \lambda_t\mathbf{1},
    \qquad
    \lambda_t
    :=
    \frac{\|h_t^{\mathrm{G}}\|_1}{q}
    =
    \frac{B\|c_t\|_2^2}{q},
    \label{eq:damped-target}
\end{equation}
where $\mathbf{1}\in\mathbb{R}^d$ is the all-ones vector.
This is a Levenberg--Marquardt-type damping of the GNB diagonal, of the
kind commonly added to curvature preconditioners for numerical
stability. Here it is not a tuned hyperparameter but is induced by the
finite budget, and $\tilde h_t\rightarrow h_t^{\mathrm{G}}$ as
$q\rightarrow\infty$. Adding a constant to every coordinate preserves
the ordering of the coordinates by curvature.

The reference is coupled to the ZO-Curv trajectory: at every iteration,
$\tilde h_t$ is evaluated at the ZO-Curv iterate $\theta_t$, on the
same mini-batch and with the same sampled labels as the ZO queries.
It is never computed by the algorithm and serves only as a comparison
point. Define the reference curvature EMA, initialized at zero as in
ZO-Curv, as
\begin{equation}
    \tilde v_t
    =
    \beta_2 \tilde v_{t-1}
    +
    (1-\beta_2)\tilde h_t,
    \qquad
    \tilde v_0=0.
    \label{eq:oracle-ema}
\end{equation}

The corresponding reference denominator is
\begin{equation}
    d_{t,i}
    :=
    \max\{
        \gamma \tilde v_{t,i},
        \epsilon_0
    \},
    \qquad
    D_t:=\operatorname{Diag}(d_t),
    \label{eq:reference-denominator}
\end{equation}
where $\epsilon_0>0$ is the numerical denominator floor.
We use $\varepsilon$ exclusively for the ZO finite-difference radius,
to distinguish it from $\epsilon_0$.

\subsection{ZO approximation of the GNB curvature}
\label{app:proofs:estimator}

At each iteration, ZO-Curv draws
$q$ independent Gaussian perturbations
\begin{equation}
    u_{t,j}
    \overset{\mathrm{i.i.d.}}{\sim}
    \mathcal{N}(0,I_d),
    \qquad
    j=1,\ldots,q.
\end{equation}
Importantly, all $q$ queries use the same mini-batch and the same sampled
pseudo-labels, and hence share the same curvature gradient $c_t$.

For direction $u_{t,j}$, define the central-difference directional
derivative
\begin{equation}
    \Delta_{t,j}^{\varepsilon}
    :=
    \frac{
        \ell_t^{\mathrm{samp}}
        (\theta_t+\varepsilon u_{t,j})
        -
        \ell_t^{\mathrm{samp}}
        (\theta_t-\varepsilon u_{t,j})
    }{
        2\varepsilon
    }.
    \label{eq:central-difference}
\end{equation}
The corresponding ZO gradient estimate is
\begin{equation}
    z_{t,j}^{\varepsilon}
    :=
    \Delta_{t,j}^{\varepsilon}u_{t,j}.
\end{equation}
ZO-Curv first averages the $q$ gradient estimates,
\begin{equation}
    \bar z_t^{\varepsilon}
    :=
    \frac{1}{q}
    \sum_{j=1}^{q}
    z_{t,j}^{\varepsilon},
    \label{eq:zo-average}
\end{equation}
and then squares the averaged estimate:
\begin{equation}
    h_t^{\mathrm{ZO}}
    :=
    B\,
    \bar z_t^{\varepsilon}
    \odot
    \bar z_t^{\varepsilon}.
    \label{eq:zo-curvature}
\end{equation}

We first study the ideal directional-derivative limit
$\varepsilon\rightarrow0$.

\begin{lemma}[Moments of the averaged ZO gradient]
\label{lem:zo-moments}
Fix a vector $c\in\mathbb{R}^{d}$ and let
$u_1,\ldots,u_q\overset{\mathrm{i.i.d.}}{\sim}\mathcal{N}(0,I_d)$.
Define
\begin{equation}
    z_j
    :=
    (c^\top u_j)u_j,
    \qquad
    \bar z
    :=
    \frac{1}{q}\sum_{j=1}^{q}z_j.
\end{equation}
Then, for every coordinate $i$,
\begin{align}
    \mathbb{E}[\bar z_i\mid c]
    &=
    c_i,
    \label{eq:zo-first-moment}\\
    \mathbb{E}[\bar z_i^2\mid c]
    &=
    \left(1+\frac{1}{q}\right)c_i^2
    +
    \frac{1}{q}\|c\|_2^2,
    \label{eq:zo-second-moment}\\
    \operatorname{Var}(\bar z_i\mid c)
    &=
    \frac{
        \|c\|_2^2+c_i^2
    }{q}.
    \label{eq:zo-coordinate-variance}
\end{align}
\end{lemma}

\emph{Proof.}
For a single Gaussian direction,
\begin{equation}
    \mathbb{E}[
        (c^\top u)u_i
    ]
    =
    \sum_{r=1}^{d}
    c_r\mathbb{E}[u_ru_i]
    =
    c_i,
\end{equation}
which proves \eqref{eq:zo-first-moment}.

For the second moment, Isserlis' identity for a standard Gaussian gives
\begin{align}
    \mathbb{E}
    \left[
        (c^\top u)^2u_i^2
    \right]
    &=
    \|c\|_2^2+2c_i^2.
\end{align}
Therefore
\begin{equation}
    \operatorname{Var}
    \left(
        (c^\top u)u_i
    \right)
    =
    \|c\|_2^2+c_i^2.
\end{equation}
Averaging $q$ independent directions divides the variance by $q$, hence
\begin{equation}
    \operatorname{Var}(\bar z_i\mid c)
    =
    \frac{
        \|c\|_2^2+c_i^2
    }{q}.
\end{equation}
Combining this with
$\mathbb{E}[\bar z_i\mid c]=c_i$ yields
\begin{align}
    \mathbb{E}[\bar z_i^2\mid c]
    &=
    c_i^2
    +
    \frac{
        \|c\|_2^2+c_i^2
    }{q}\\
    &=
    \left(1+\frac{1}{q}\right)c_i^2
    +
    \frac{1}{q}\|c\|_2^2.
\end{align}
\hfill$\square$

Lemma~\ref{lem:zo-moments} shows that the averaged ZO gradient itself
approaches $c$ at the canonical $q^{-1/2}$ stochastic rate.
Its second moment~\eqref{eq:zo-second-moment} is the source of the
damping in~\eqref{eq:damped-target}.
For the convergence analysis, however, we require a high-probability
rather than an expectation bound.

\begin{lemma}[High-probability ZO gradient error]
\label{lem:zo-concentration}
Under the setting of Lemma~\ref{lem:zo-moments}, there exists a universal
constant $C_0>0$ such that, for any failure probability
$\zeta\in(0,1)$, with probability at least $1-\zeta$,
\begin{equation}
    \|\bar z-c\|_\infty
    \le
    C_0\|c\|_2
    \left[
        \sqrt{
            \frac{
                \log(2d/\zeta)
            }{q}
        }
        +
        \frac{
            \log(2d/\zeta)
        }{q}
    \right].
    \label{eq:zo-gradient-concentration}
\end{equation}
\end{lemma}

\emph{Proof.}
Fix a coordinate $i$ and define
\begin{equation}
    X_{j,i}
    :=
    (c^\top u_j)u_{j,i}-c_i.
\end{equation}
We have
$\mathbb{E}[X_{j,i}]=0$.
Moreover,
$c^\top u_j$ is a centered Gaussian with variance $\|c\|_2^2$,
and $u_{j,i}$ is standard Gaussian.
The product of two sub-Gaussian random variables is sub-exponential,
even when the two variables are correlated.
Therefore, for a universal constant $C>0$,
\begin{equation}
    \|X_{j,i}\|_{\psi_1}
    \le
    C\|c\|_2.
\end{equation}
Bernstein's inequality for independent centered sub-exponential
variables then gives
\begin{equation}
    \Pr\left[
        \left|
            \frac{1}{q}
            \sum_{j=1}^{q}X_{j,i}
        \right|
        >
        C_0\|c\|_2
        \left(
            \sqrt{\frac{s}{q}}
            +
            \frac{s}{q}
        \right)
    \right]
    \le
    2e^{-s}.
\end{equation}
Setting
\begin{equation}
    s=\log(2d/\zeta)
\end{equation}
and applying a union bound over $i=1,\ldots,d$ proves
\eqref{eq:zo-gradient-concentration}.
\hfill$\square$

We next account for the finite central-difference radius
$\varepsilon$.

\begin{assumption}[Local Hessian Lipschitzness]
\label{ass:hessian-lipschitz}
For every iteration $t$, the sampled-label loss
$\ell_t^{\mathrm{samp}}$ is twice differentiable with an
$L_H$-Lipschitz Hessian on the ball
$\mathcal U_t:=\{\theta:\|\theta-\theta_t\|_2\le\varepsilon R\}$
containing all ZO query points:
\begin{equation}
    \|
        \nabla^2\ell_t^{\mathrm{samp}}(\theta)
        -
        \nabla^2\ell_t^{\mathrm{samp}}(\theta')
    \|_2
    \le
    L_H\|\theta-\theta'\|_2,
    \qquad
    \theta,\theta'\in\mathcal U_t.
    \label{eq:hessian-lipschitz}
\end{equation}
\end{assumption}
The query points are $\theta_t\pm\varepsilon u_{t,j}$ with
$\|u_{t,j}\|_2\approx\sqrt d$, so the relevant radius is
$\varepsilon\max_j\|u_{t,j}\|_2$ rather than $\varepsilon$; we take
$R=R_q$ in Proposition~\ref{prop:main-curvature} and $R=R_T$ in
Theorem~\ref{thm:query-dependent}, where $R_q$ and $R_T$ bound
$\max_j\|u_{t,j}\|_2$ with high probability.

\begin{lemma}[Central-difference error]
\label{lem:fd-error}
Under Assumption~\ref{ass:hessian-lipschitz}, for any direction $u$
with $\|u\|_2\le R$,
\begin{equation}
    \left|
        \frac{
            \ell_t^{\mathrm{samp}}(\theta_t+\varepsilon u)
            -
            \ell_t^{\mathrm{samp}}(\theta_t-\varepsilon u)
        }{
            2\varepsilon
        }
        -
        c_t^\top u
    \right|
    \le
    \frac{
        L_H\varepsilon^2
    }{6}
    \|u\|_2^3.
    \label{eq:central-difference-error}
\end{equation}
\end{lemma}

\emph{Proof.}
Define the scalar function
\begin{equation}
    \phi(s)
    :=
    \ell_t^{\mathrm{samp}}(\theta_t+su),
    \qquad
    |s|\le\varepsilon.
\end{equation}
Then $\phi'(0)=c_t^\top u$ and $\phi''(s)=u^\top\nabla^2
\ell_t^{\mathrm{samp}}(\theta_t+su)u$, so
Assumption~\ref{ass:hessian-lipschitz} gives
\begin{equation}
    |\phi''(s)-\phi''(0)|
    \le
    L_H|s|\|u\|_2^3.
\end{equation}
Taylor's theorem with integral remainder,
$\phi(\pm\varepsilon)=\phi(0)\pm\varepsilon\phi'(0)
+\int_0^{\pm\varepsilon}(\pm\varepsilon-s)\phi''(s)\,ds$,
then yields
\begin{equation}
    \left|
        \phi(\pm\varepsilon)-\phi(0)\mp\varepsilon\phi'(0)
        -\frac{\varepsilon^2}{2}\phi''(0)
    \right|
    \le
    L_H\|u\|_2^3
    \int_0^{\varepsilon}(\varepsilon-s)s\,ds
    =
    \frac{L_H\varepsilon^3}{6}\|u\|_2^3.
\end{equation}
Subtracting the two expansions cancels $\phi(0)$ and $\phi''(0)$, and
dividing by $2\varepsilon$ gives
\begin{equation}
    \left|
        \frac{
            \phi(\varepsilon)-\phi(-\varepsilon)
        }{
            2\varepsilon
        }
        -
        \phi'(0)
    \right|
    \le
    \frac{
        L_H\varepsilon^2
    }{6}
    \|u\|_2^3.
\end{equation}
This proves the result; no third derivative is required.
\hfill$\square$

We now combine Lemmas~\ref{lem:zo-moments}--\ref{lem:fd-error} into the
properties of the curvature estimate used in the rest of the analysis.
Write
\begin{equation}
    h_t^{\mathrm{ZO},0}
    :=
    B\,
    \bar z_t^{0}
    \odot
    \bar z_t^{0},
    \qquad
    \bar z_t^{0}
    :=
    \frac{1}{q}
    \sum_{j=1}^{q}
    (c_t^\top u_{t,j})u_{t,j},
    \label{eq:ideal-zo-curvature}
\end{equation}
for the ideal estimate with exact directional derivatives, and
$h_t^{\mathrm{ZO}}$ for the central-difference
estimate~\eqref{eq:zo-curvature}.

\begin{restated}{Proposition}{prop:main-curvature}[Forward-only estimation of damped GNB curvature]
Fix an iteration $t$ and condition on its mini-batch and sampled labels,
with $c_t\neq0$.
\begin{enumerate}
    \item[(i)] \emph{Unbiasedness.} The ideal estimate is unbiased for
    the damped target~\eqref{eq:damped-target}:
    \begin{equation}
        \mathbb{E}\bigl[h_t^{\mathrm{ZO},0}\bigr]
        =
        \tilde h_t.
        \label{eq:zo-unbiased}
    \end{equation}
    \item[(ii)] \emph{Relative second moment.} There is a universal
    constant $C_v\ge1$ such that, for every coordinate $i$,
    \begin{equation}
        \mathbb{E}\bigl[(h_{t,i}^{\mathrm{ZO},0})^2\bigr]
        \le
        C_v\tilde h_{t,i}^2.
        \label{eq:zo-relative-moment}
    \end{equation}
    \item[(iii)] \emph{Relative envelope and finite differences.} For
    $\zeta\in(0,1)$, define
    \begin{equation}
    \begin{gathered}
        \Lambda
        :=
        \log\frac{4d}{\zeta},
        \qquad
        R_q
        :=
        \sqrt{d}
        +
        \sqrt{2\log\frac{4q}{\zeta}},
        \\
        M_{q,d}
        :=
        \sqrt{2\log\frac{8qd}{\zeta}},
        \qquad
        \vartheta
        :=
        \frac{L_H\varepsilon^2}{6}
        R_q^3M_{q,d},
    \end{gathered}
        \label{eq:prop-constants}
    \end{equation}
    and suppose $q\ge\Lambda$ and that
    Assumption~\ref{ass:hessian-lipschitz} holds with $R=R_q$. Then,
    with probability at least $1-\zeta$, simultaneously for all
    coordinates $i$,
    \begin{align}
        h_{t,i}^{\mathrm{ZO},0}
        &\le
        C_b\Lambda\,\tilde h_{t,i},
        \qquad
        C_b:=\max\{2,8C_0^2\},
        \label{eq:zo-envelope}\\
        \bigl|
            h_{t,i}^{\mathrm{ZO}}
            -
            h_{t,i}^{\mathrm{ZO},0}
        \bigr|
        &\le
        2\sqrt{C_b\Lambda B\vartheta^2\,\tilde h_{t,i}}
        +
        B\vartheta^2,
        \label{eq:zo-fd-perturbation}
    \end{align}
    where $C_0$ is the constant of Lemma~\ref{lem:zo-concentration}.
\end{enumerate}
\end{restated}

\emph{Proof.}
For (i), Lemma~\ref{lem:zo-moments} with $c=c_t$ gives
\begin{equation}
    \mathbb{E}\bigl[h_{t,i}^{\mathrm{ZO},0}\bigr]
    =
    B\,\mathbb{E}\bigl[(\bar z_{t,i}^{0})^2\bigr]
    =
    \left(1+\frac{1}{q}\right)Bc_{t,i}^2
    +
    \frac{B\|c_t\|_2^2}{q}
    =
    \tilde h_{t,i}.
\end{equation}

For (ii), write $\bar z_{t,i}^{0}=c_{t,i}+a_i$ with
$a_i:=q^{-1}\sum_{j=1}^{q}X_{j,i}$ and $X_{j,i}$ as in the proof of
Lemma~\ref{lem:zo-concentration}. Since $(x+y)^4\le8(x^4+y^4)$,
\begin{equation}
    \mathbb{E}\bigl[(\bar z_{t,i}^{0})^4\bigr]
    \le
    8c_{t,i}^4
    +
    8\,\mathbb{E}[a_i^4].
\end{equation}
The $X_{j,i}$ are independent and centered, so
\begin{equation}
    \mathbb{E}[a_i^4]
    =
    \frac{
        q\,\mathbb{E}[X_{1,i}^4]
        +
        3q(q-1)\bigl(\mathbb{E}[X_{1,i}^2]\bigr)^2
    }{q^4}
    \le
    \frac{
        \mathbb{E}[X_{1,i}^4]
        +
        3\bigl(\mathbb{E}[X_{1,i}^2]\bigr)^2
    }{q^2}.
\end{equation}
By Lemma~\ref{lem:zo-moments},
$\mathbb{E}[X_{1,i}^2]=\|c_t\|_2^2+c_{t,i}^2\le2\|c_t\|_2^2$, and since
$\|X_{1,i}\|_{\psi_1}\le C\|c_t\|_2$, also
$\mathbb{E}[X_{1,i}^4]\le C'\|c_t\|_2^4$ for a universal constant $C'$.
Hence $\mathbb{E}[a_i^4]\le C''\|c_t\|_2^4/q^2$ with $C'':=C'+12$, and
\begin{align}
    \mathbb{E}\bigl[(h_{t,i}^{\mathrm{ZO},0})^2\bigr]
    &=
    B^2\,\mathbb{E}\bigl[(\bar z_{t,i}^{0})^4\bigr]
    \le
    8\bigl(Bc_{t,i}^2\bigr)^2
    +
    8C''\lambda_t^2\\
    &\le
    C_v\bigl(Bc_{t,i}^2+\lambda_t\bigr)^2
    \le
    C_v\tilde h_{t,i}^2,
\end{align}
with $C_v:=8\max\{1,C''\}$, where the last step uses
$Bc_{t,i}^2+\lambda_t\le\tilde h_{t,i}$.

For (iii), we split the failure probability $\zeta$ into three events.
First, Lemma~\ref{lem:zo-concentration} with failure probability
$\zeta/2$, for which $\log(2d/(\zeta/2))=\Lambda$, gives
$|a_i|\le C_0\|c_t\|_2(\sqrt{\Lambda/q}+\Lambda/q)$ for all $i$.
Since $q\ge\Lambda$ and $(\sqrt x+x)^2\le4x$ for $x\le1$,
\begin{equation}
    Ba_i^2
    \le
    4C_0^2\Lambda\frac{B\|c_t\|_2^2}{q}
    =
    4C_0^2\Lambda\lambda_t.
\end{equation}
Therefore, using $\Lambda\ge1$,
\begin{equation}
    h_{t,i}^{\mathrm{ZO},0}
    =
    B(c_{t,i}+a_i)^2
    \le
    2Bc_{t,i}^2
    +
    8C_0^2\Lambda\lambda_t
    \le
    C_b\Lambda\,\tilde h_{t,i},
\end{equation}
which is~\eqref{eq:zo-envelope}.

Second, Gaussian norm concentration,
$\Pr[\|u\|_2\ge\sqrt d+s]\le e^{-s^2/2}$, and a union bound over the $q$
directions give, with probability at least $1-\zeta/4$,
$\max_{j\le q}\|u_{t,j}\|_2\le R_q$. Third, the Gaussian tail bound
$\Pr[|u_{j,i}|\ge s]\le2e^{-s^2/2}$ and a union bound over all $qd$
coordinates give, with probability at least $1-\zeta/4$,
$\max_{j\le q,\,i\le d}|u_{t,j,i}|\le M_{q,d}$.
On the second event every query point lies in $\mathcal U_t$ with
$R=R_q$, so Lemma~\ref{lem:fd-error} applies to every direction and
\begin{equation}
    \bar z_t^{\varepsilon}
    -
    \bar z_t^{0}
    =
    \frac{1}{q}
    \sum_{j=1}^{q}
    e_{t,j}^{\mathrm{FD}},
    \qquad
    |e_{t,j,i}^{\mathrm{FD}}|
    \le
    \frac{L_H\varepsilon^2}{6}
    \|u_{t,j}\|_2^3|u_{t,j,i}|.
\end{equation}
On both events, $e_i:=\bar z_{t,i}^{\varepsilon}-\bar z_{t,i}^{0}$
satisfies $|e_i|\le\vartheta$. Then
\begin{equation}
    \bigl|
        h_{t,i}^{\mathrm{ZO}}
        -
        h_{t,i}^{\mathrm{ZO},0}
    \bigr|
    =
    B\bigl|2\bar z_{t,i}^{0}e_i+e_i^2\bigr|
    \le
    2\sqrt{B\,h_{t,i}^{\mathrm{ZO},0}}\,\vartheta
    +
    B\vartheta^2,
\end{equation}
and~\eqref{eq:zo-envelope} gives~\eqref{eq:zo-fd-perturbation}.
A union bound over the three events, with total failure probability
$\zeta/2+\zeta/4+\zeta/4=\zeta$, completes the proof.
\hfill$\square$

Proposition~\ref{prop:main-curvature} shows that, like the GNB
estimator of Sophia-G, the forward-only estimate is unbiased; its target
is the GNB diagonal with the additional uniform damping
$\lambda_t=\|h_t^{\mathrm{G}}\|_1/q$. Averaging before squaring is what
keeps this damping small: squaring each directional estimate before
averaging would give
$\mathbb{E}[B(\Delta_{t,j}^{0}u_{t,j,i})^2]=2h_{t,i}^{\mathrm{G}}
+\|h_t^{\mathrm{G}}\|_1$, a damping $q$ times larger that does not
decrease with the budget. Moreover, by~(ii) and~(iii), the fluctuation of
the estimate around its target is at most of the order of the target
itself in every coordinate, with constants independent of $d$, $q$, and
$c_t$. This relative control is what the curvature EMA exploits next.
Dividing~\eqref{eq:zo-fd-perturbation} by $\tilde h_{t,i}$ shows that
the finite-difference perturbation is also relative, of order
$\sqrt{\Lambda B\vartheta^2/\tilde h_{t,i}}=O(\varepsilon^2)$.

\subsection{Effect of curvature EMA}
\label{app:proofs:ema}

The optimizer does not use a single raw curvature estimate directly.
Instead, both Sophia-G and ZO-Curv maintain a curvature EMA. Because
fresh directions are drawn at every refresh, the EMA averages the
relative fluctuations of Proposition~\ref{prop:main-curvature} over its
effective window $1/(1-\beta_2)$.

At every iteration, define
\begin{equation}
    v_t^{\mathrm{ZO}}
    =
    \beta_2v_{t-1}^{\mathrm{ZO}}
    +
    (1-\beta_2)h_t^{\mathrm{ZO}},
    \qquad
    v_0^{\mathrm{ZO}}=0,
    \label{eq:zo-ema}
\end{equation}
and let $w_{t,s}:=(1-\beta_2)\beta_2^{t-s}$ and
$W_t:=\sum_{s=1}^{t}w_{t,s}=1-\beta_2^{t}$, so that
$v_t^{\mathrm{ZO}}=\sum_{s\le t}w_{t,s}h_s^{\mathrm{ZO}}$ and
$\tilde v_t=\sum_{s\le t}w_{t,s}\tilde h_s$.
Let $\mathcal G_s$ denote the history up to and including $\theta_s$ and
the step-$s$ mini-batch and sampled labels, but not the step-$s$
directions. Then $c_s$ and $\tilde h_s$ are $\mathcal G_s$-measurable,
the directions $u_{s,1},\ldots,u_{s,q}$ are independent of
$\mathcal G_s$, and $h_s^{\mathrm{ZO}}$ is
$\mathcal G_{s+1}$-measurable.

\begin{assumption}[Slowly varying damped curvature]
\label{ass:slow-curvature}
Over the analyzed horizon $s\le T$, there exist $\kappa\ge1$ and
deterministic $\bar H_i>0$ such that
\begin{equation}
    \frac{\bar H_i}{\kappa}
    \le
    \tilde h_{s,i}
    \le
    \bar H_i,
    \qquad
    s\le T,\;
    i=1,\ldots,d.
    \label{eq:slow-curvature}
\end{equation}
almost surely along the ZO-Curv trajectory.
\end{assumption}
Since $\tilde h_{s,i}\ge\lambda_s$, the damping keeps every coordinate
away from zero, and the assumption only asks that the damped curvature
of each coordinate varies by at most a factor $\kappa$ over the horizon.
Let $\underline H:=\min_i\bar H_i$.

\begin{lemma}[EMA concentration]
\label{lem:ema-error}
Suppose Assumptions~\ref{ass:hessian-lipschitz}
and~\ref{ass:slow-curvature} hold over $T$ iterations. For
$\zeta\in(0,1)$, define
\begin{equation}
\begin{gathered}
    \Lambda_T
    :=
    \log\frac{16qdT}{\zeta},
    \qquad
    R_T
    :=
    \sqrt{d}
    +
    \sqrt{2\log\frac{8qT}{\zeta}},
    \\
    \vartheta_T
    :=
    \frac{L_H\varepsilon^2}{6}
    R_T^3\sqrt{2\Lambda_T},
    \qquad
    \varrho_T
    :=
    \frac{\kappa B\vartheta_T^2}{\underline H},
\end{gathered}
    \label{eq:uniform-constants}
\end{equation}
take $R=R_T$ in Assumption~\ref{ass:hessian-lipschitz}, and suppose
$q\ge\Lambda_T$. Let $t_0:=\lceil\ln2/(1-\beta_2)\rceil$ and
\begin{equation}
    \delta_T
    :=
    2\kappa
    \left[
        \sqrt{2C_v(1-\beta_2)\Lambda_T}
        +
        \frac{4}{3}C_b(1-\beta_2)\Lambda_T^2
    \right]
    +
    \sqrt{\frac{C_v\zeta}{2T}}
    +
    2\sqrt{C_b\Lambda_T\varrho_T}
    +
    \varrho_T.
    \label{eq:uniform-delta}
\end{equation}
Then, with probability at least $1-\zeta$, simultaneously for all
$t_0\le t\le T$ and all coordinates $i$,
\begin{equation}
    \bigl|
        v_{t,i}^{\mathrm{ZO}}
        -
        \tilde v_{t,i}
    \bigr|
    \le
    \delta_T\,\tilde v_{t,i}.
    \label{eq:ema-error-bound}
\end{equation}
\end{lemma}

\emph{Proof.}
For each $s\le T$, let $\mathcal E_s$ be the event of
Proposition~\ref{prop:main-curvature}(iii) at step $s$ with failure
probability $\zeta/(2T)$. Its constants satisfy
$\log(8dT/\zeta)\le\Lambda_T$, $R_q=R_T$, and
$M_{q,d}=\sqrt{2\log(16qdT/\zeta)}=\sqrt{2\Lambda_T}$, so the bounds of
Proposition~\ref{prop:main-curvature}(iii) hold with $\Lambda_T$ and
$\vartheta_T$. Because the step-$s$ directions are independent of
$\mathcal G_s$, the proposition applies conditionally on $\mathcal G_s$,
and $\Pr[\mathcal E_s^c\mid\mathcal G_s]\le\zeta/(2T)$.

Fix a coordinate $i$ and decompose
\begin{equation}
    v_{t,i}^{\mathrm{ZO}}
    -
    \tilde v_{t,i}
    =
    \sum_{s\le t}
    w_{t,s}
    \bigl(
        h_{s,i}^{\mathrm{ZO}}
        -
        h_{s,i}^{\mathrm{ZO},0}
    \bigr)
    +
    \sum_{s\le t}
    w_{t,s}\,\xi_{s,i},
    \qquad
    \xi_{s,i}
    :=
    h_{s,i}^{\mathrm{ZO},0}
    -
    \tilde h_{s,i}.
\end{equation}

\emph{Finite-difference term.} On $\mathcal E_s$,
\eqref{eq:zo-fd-perturbation} and
$\tilde h_{s,i}\ge\bar H_i/\kappa\ge\underline H/\kappa$ give
\begin{equation}
    \bigl|
        h_{s,i}^{\mathrm{ZO}}
        -
        h_{s,i}^{\mathrm{ZO},0}
    \bigr|
    \le
    \tilde h_{s,i}
    \left[
        2\sqrt{\frac{C_b\Lambda_TB\vartheta_T^2}{\tilde h_{s,i}}}
        +
        \frac{B\vartheta_T^2}{\tilde h_{s,i}}
    \right]
    \le
    \bigl(
        2\sqrt{C_b\Lambda_T\varrho_T}
        +
        \varrho_T
    \bigr)
    \tilde h_{s,i},
\end{equation}
so the first sum is at most
$(2\sqrt{C_b\Lambda_T\varrho_T}+\varrho_T)\tilde v_{t,i}$.

\emph{Truncation.} By Proposition~\ref{prop:main-curvature}(i),
$\mathbb{E}[\xi_{s,i}\mid\mathcal G_s]=0$. Let
$b_i:=C_b\Lambda_T\bar H_i$. Since $h_{s,i}^{\mathrm{ZO},0}$ and
$\tilde h_{s,i}$ are nonnegative and $C_b\Lambda_T\ge1$, on
$\mathcal E_s$ we have
$|\xi_{s,i}|\le C_b\Lambda_T\tilde h_{s,i}\le b_i$. Define the truncated
martingale differences
\begin{equation}
    \xi_{s,i}'
    :=
    \xi_{s,i}\mathbb{1}\{|\xi_{s,i}|\le b_i\}
    -
    \mathbb{E}\bigl[
        \xi_{s,i}\mathbb{1}\{|\xi_{s,i}|\le b_i\}
        \bigm|
        \mathcal G_s
    \bigr].
\end{equation}
Because $\mathbb{E}[\xi_{s,i}\mid\mathcal G_s]=0$, the Cauchy--Schwarz
inequality and~\eqref{eq:zo-relative-moment} bound the conditional mean
that is removed:
\begin{equation}
    \Bigl|
        \mathbb{E}\bigl[
            \xi_{s,i}\mathbb{1}\{|\xi_{s,i}|\le b_i\}
            \bigm|
            \mathcal G_s
        \bigr]
    \Bigr|
    =
    \Bigl|
        \mathbb{E}\bigl[
            \xi_{s,i}\mathbb{1}\{|\xi_{s,i}|>b_i\}
            \bigm|
            \mathcal G_s
        \bigr]
    \Bigr|
    \le
    \sqrt{C_v}\,\tilde h_{s,i}
    \sqrt{\frac{\zeta}{2T}}.
\end{equation}
Hence, on $\bigcap_{s\le T}\mathcal E_s$,
\begin{equation}
    \Bigl|
        \sum_{s\le t}w_{t,s}\xi_{s,i}
    \Bigr|
    \le
    \Bigl|
        \sum_{s\le t}w_{t,s}\xi_{s,i}'
    \Bigr|
    +
    \sqrt{\frac{C_v\zeta}{2T}}\,\tilde v_{t,i}.
\end{equation}

\emph{Martingale term.} The sequence $Y_s:=w_{t,s}\xi_{s,i}'$ is a
martingale difference sequence with respect to
$(\mathcal G_{s+1})_s$, with $|Y_s|\le2(1-\beta_2)b_i$ and, by
Assumption~\ref{ass:slow-curvature} and~\eqref{eq:zo-relative-moment},
\begin{equation}
    \sum_{s\le t}
    \mathbb{E}\bigl[Y_s^2\mid\mathcal G_s\bigr]
    \le
    C_v\bar H_i^2
    \sum_{s\le t}w_{t,s}^2
    \le
    C_v(1-\beta_2)\bar H_i^2.
\end{equation}
Freedman's inequality then gives, with probability at least
$1-\zeta/(2dT)$,
\begin{equation}
    \Bigl|
        \sum_{s\le t}Y_s
    \Bigr|
    \le
    \bar H_i
    \left[
        \sqrt{2C_v(1-\beta_2)\Lambda_T}
        +
        \frac{4}{3}C_b(1-\beta_2)\Lambda_T^2
    \right],
\end{equation}
where we used $\log(4dT/\zeta)\le\Lambda_T$.
Finally, for $t\ge t_0$ we have $\beta_2^{t}\le e^{-(1-\beta_2)t}\le1/2$,
so $W_t\ge1/2$ and
$\tilde v_{t,i}\ge W_t\bar H_i/\kappa\ge\bar H_i/(2\kappa)$ by
Assumption~\ref{ass:slow-curvature}. Dividing by $\tilde v_{t,i}$ and
collecting the three terms gives~\eqref{eq:ema-error-bound} for this
pair $(t,i)$. A union bound over the events $\mathcal E_s$, with total
failure probability $\zeta/2$, and over the $dT$ Freedman events, with
total failure probability $\zeta/2$, completes the proof.
\hfill$\square$

Lemma~\ref{lem:ema-error} bounds the error relative to the reference
EMA itself. The leading term is governed by the EMA window
$1/(1-\beta_2)$ and depends on $d$, $q$, and $T$ only through
$\Lambda_T$. It does not require $q\rightarrow\infty$: the query budget
enters the analysis through the damping $\lambda_t$ of the target
rather than through the estimation error.

\subsection{From curvature error to preconditioner distortion}
\label{app:proofs:distortion}

Define the actual ZO-Curv denominator
\begin{equation}
    \hat d_{t,i}
    :=
    \max\{
        \gamma v_{t,i}^{\mathrm{ZO}},
        \epsilon_0
    \},
    \qquad
    \widehat D_t
    :=
    \operatorname{Diag}(\hat d_t).
    \label{eq:zo-denominator}
\end{equation}

\begin{lemma}[Denominator distortion]
\label{lem:denominator-distortion}
Suppose that, for some $\delta\in[0,1)$ and every coordinate $i$,
\begin{equation}
    (1-\delta)\tilde v_{t,i}
    \le
    v_{t,i}^{\mathrm{ZO}}
    \le
    (1+\delta)\tilde v_{t,i}.
\end{equation}
Then
\begin{equation}
    (1-\delta)D_t
    \preceq
    \widehat D_t
    \preceq
    (1+\delta)D_t.
    \label{eq:multiplicative-distortion}
\end{equation}
\end{lemma}

\emph{Proof.}
The map $x\mapsto\max\{\gamma x,\epsilon_0\}$ is nondecreasing. For the
upper bound,
\begin{equation}
    \hat d_{t,i}
    \le
    \max\{
        (1+\delta)\gamma\tilde v_{t,i},
        \epsilon_0
    \}
    \le
    (1+\delta)
    \max\{
        \gamma\tilde v_{t,i},
        \epsilon_0
    \}
    =
    (1+\delta)d_{t,i}.
\end{equation}
For the lower bound, since $\epsilon_0\ge(1-\delta)\epsilon_0$,
\begin{equation}
    \hat d_{t,i}
    \ge
    \max\{
        (1-\delta)\gamma\tilde v_{t,i},
        \epsilon_0
    \}
    \ge
    (1-\delta)
    \max\{
        \gamma\tilde v_{t,i},
        \epsilon_0
    \}
    =
    (1-\delta)d_{t,i}.
\end{equation}
Both matrices are diagonal, which proves
\eqref{eq:multiplicative-distortion}.
\hfill$\square$

Lemma~\ref{lem:denominator-distortion} is the interface between the
estimator analysis and the optimization analysis.
Because the error bound of Lemma~\ref{lem:ema-error} is relative, it
passes through the floor without any division by $\epsilon_0$, and the
entire effect of the forward-only ZO estimator is summarized by the
multiplicative distortion $\delta$.

\subsection{Robust descent under approximate curvature}
\label{app:proofs:robust-descent}

We now analyze the deterministic clipped preconditioned update
\begin{equation}
    \theta_{t+1}
    =
    \theta_t-\eta s_t,
    \label{eq:analysis-update}
\end{equation}
where
\begin{equation}
    s_t
    :=
    \operatorname{clip}
    \left(
        \widehat D_t^{-1}g_t,
        \rho
    \right).
    \label{eq:analysis-direction}
\end{equation}
The clipping operator is applied coordinate-wise:
\begin{equation}
    \operatorname{clip}(x,\rho)
    :=
    \operatorname{sign}(x)
    \min\{|x|,\rho\}.
\end{equation}
The implementation corresponds to $\rho=1$.

The following assumption is a preconditioned analogue of the usual
smoothness assumption.

\begin{assumption}[Local preconditioned smoothness]
\label{ass:preconditioned-smoothness}
There exists $\alpha>0$ such that, for every iteration $t$ and every
point on the update segment,
\begin{equation}
    \nabla^2F(
        \theta_t-\tau\eta s_t
    )
    \preceq
    \alpha D_t,
    \qquad
    \forall\tau\in[0,1].
    \label{eq:preconditioned-smoothness}
\end{equation}
\end{assumption}

For ordinary gradient descent, smoothness assumes
$\nabla^2F\preceq LI$.
Assumption~\ref{ass:preconditioned-smoothness} replaces the isotropic
matrix $LI$ by the reference curvature preconditioner $D_t$.
Because the reference target is damped, ordinary smoothness already
implies this assumption. For $t\ge t_0$ (defined in
Lemma~\ref{lem:ema-error}),
$d_{t,i}\ge\gamma\tilde v_{t,i}\ge\gamma W_t\min_{s\le t}\lambda_s
\ge\frac{\gamma}{2}\min_{s\le t}\lambda_s$, so $\nabla^2F\preceq LI$
along the update segments implies
Assumption~\ref{ass:preconditioned-smoothness} with
$\alpha=2L/(\gamma\min_{s\le t}\lambda_s)$. This is only a sufficient
condition; the assumption holds with a smaller $\alpha$ whenever the
GNB part of $D_t$ captures the curvature of $F$.

\begin{restated}{Theorem}{thm:main-descent}[Robust descent under curvature estimation error]
Suppose
\begin{equation}
    (1-\delta)D_t
    \preceq
    \widehat D_t
    \preceq
    (1+\delta)D_t
    \label{eq:thm-distortion}
\end{equation}
for some $\delta\in[0,1)$, and
Assumption~\ref{ass:preconditioned-smoothness} holds.
Define
\begin{equation}
    \Psi_t
    :=
    g_t^\top s_t.
    \label{eq:stationarity-measure}
\end{equation}
Then
\begin{equation}
    \Psi_t
    =
    \sum_{i=1}^{d}
    \min
    \left\{
        \frac{g_{t,i}^2}{\hat d_{t,i}},
        \rho|g_{t,i}|
    \right\}
    \ge0,
    \label{eq:psi-expanded}
\end{equation}
and the one-step loss satisfies
\begin{equation}
    F(\theta_{t+1})
    \le
    F(\theta_t)
    -
    \eta
    \left(
        1-
        \frac{
            \alpha\eta
        }{
            2(1-\delta)
        }
    \right)
    \Psi_t.
    \label{eq:robust-descent-bound}
\end{equation}
Therefore, every non-stationary step is a descent step whenever
\begin{equation}
    0<\eta<
    \frac{
        2(1-\delta)
    }{\alpha}.
    \label{eq:stable-stepsize}
\end{equation}
In particular, if
\begin{equation}
    \eta
    \le
    \frac{
        1-\delta
    }{\alpha},
    \label{eq:conservative-stepsize}
\end{equation}
then
\begin{equation}
    F(\theta_{t+1})
    \le
    F(\theta_t)
    -
    \frac{\eta}{2}\Psi_t.
    \label{eq:half-descent}
\end{equation}
\end{restated}

\emph{Proof.}
Since every $\hat d_{t,i}>0$, clipping preserves the sign of
$g_{t,i}$. Hence
\begin{align}
    g_{t,i}s_{t,i}
    &=
    |g_{t,i}|
    \min
    \left\{
        \frac{|g_{t,i}|}{\hat d_{t,i}},
        \rho
    \right\}\\
    &=
    \min
    \left\{
        \frac{g_{t,i}^2}{\hat d_{t,i}},
        \rho|g_{t,i}|
    \right\}
    \ge0.
\end{align}
Summing over coordinates gives
\eqref{eq:psi-expanded}.

Taylor's theorem with integral remainder and
Assumption~\ref{ass:preconditioned-smoothness} imply
\begin{align}
    F(\theta_t-\eta s_t)
    &\le
    F(\theta_t)
    -
    \eta g_t^\top s_t
    +
    \frac{
        \alpha\eta^2
    }{2}
    s_t^\top D_t s_t.
    \label{eq:taylor-preconditioned}
\end{align}

It remains to upper-bound the quadratic term.
By definition of clipping,
\begin{equation}
    |s_{t,i}|
    \le
    \frac{|g_{t,i}|}{\hat d_{t,i}}.
\end{equation}
Therefore
\begin{align}
    d_{t,i}s_{t,i}^2
    &\le
    d_{t,i}
    |s_{t,i}|
    \frac{|g_{t,i}|}{\hat d_{t,i}}\\
    &=
    \frac{
        d_{t,i}
    }{
        \hat d_{t,i}
    }
    |g_{t,i}||s_{t,i}|.
\end{align}
From
$\hat d_{t,i}\ge(1-\delta)d_{t,i}$,
\begin{equation}
    \frac{
        d_{t,i}
    }{
        \hat d_{t,i}
    }
    \le
    \frac{1}{1-\delta}.
\end{equation}
Since $g_{t,i}$ and $s_{t,i}$ have the same sign,
\begin{equation}
    |g_{t,i}||s_{t,i}|
    =
    g_{t,i}s_{t,i}.
\end{equation}
Thus
\begin{equation}
    d_{t,i}s_{t,i}^2
    \le
    \frac{
        1
    }{
        1-\delta
    }
    g_{t,i}s_{t,i}.
\end{equation}
Summing over $i$ yields
\begin{equation}
    s_t^\top D_t s_t
    \le
    \frac{
        1
    }{
        1-\delta
    }
    g_t^\top s_t
    =
    \frac{
        1
    }{
        1-\delta
    }
    \Psi_t.
    \label{eq:quadratic-control}
\end{equation}
Substituting
\eqref{eq:quadratic-control}
into
\eqref{eq:taylor-preconditioned}
gives
\eqref{eq:robust-descent-bound}.
The remaining claims follow immediately from the corresponding
stepsize choices.
\hfill$\square$

The key property in Theorem~\ref{thm:main-descent} is that curvature
error affects the magnitude of the update but not its coordinate-wise
sign. In contrast, replacing $g_t$ itself with a noisy ZO gradient may
change the sign of the first-order descent term
$g_t^\top\hat g_t$.
This difference formalizes why noisy ZO information is more naturally
tolerated as a preconditioner than as a descent direction.

\begin{corollary}[Convergence to stationarity]
\label{cor:stationarity}
Assume the conditions of Theorem~\ref{thm:main-descent},
let
\begin{equation}
    \eta
    \le
    \frac{1-\delta}{\alpha},
\end{equation}
and suppose $F$ is bounded below by $F_\star$.
Then, for any $T\ge1$,
\begin{equation}
    \min_{0\le t<T}
    \Psi_t
    \le
    \frac{
        2\left(
            F(\theta_0)-F_\star
        \right)
    }{
        \eta T
    }.
    \label{eq:stationarity-rate}
\end{equation}
Consequently,
\begin{equation}
    \min_{0\le t<T}\Psi_t
    =
    O(T^{-1}).
\end{equation}
\end{corollary}

\emph{Proof.}
Equation~\eqref{eq:half-descent} gives
\begin{equation}
    F(\theta_t)-F(\theta_{t+1})
    \ge
    \frac{\eta}{2}\Psi_t.
\end{equation}
Summing from $t=0$ to $T-1$,
\begin{equation}
    F(\theta_0)-F(\theta_T)
    \ge
    \frac{\eta}{2}
    \sum_{t=0}^{T-1}\Psi_t.
\end{equation}
Since
$F(\theta_T)\ge F_\star$,
\begin{equation}
    \sum_{t=0}^{T-1}\Psi_t
    \le
    \frac{
        2(F(\theta_0)-F_\star)
    }{\eta}.
\end{equation}
The minimum is no larger than the average, giving
\eqref{eq:stationarity-rate}.
\hfill$\square$

\begin{remark}[From $\Psi_t$ to the gradient]
\label{rem:psi-gradient}
The measure $\Psi_t$ controls the gradient whenever the denominator is
bounded above. Under Assumption~\ref{ass:slow-curvature},
$\tilde v_{t,i}\le W_t\bar H_i\le\bar H_i$, so on the event of
Theorem~\ref{thm:query-dependent},
$\hat d_{t,i}\le(1+\delta_T)d_{t,i}\le\hat d_{\max}
:=(1+\delta_T)\max\{\gamma\max_i\bar H_i,\epsilon_0\}$.
Since $a\mapsto\min\{a^2/\hat d_{\max},\rho|a|\}$ is increasing in $|a|$,
$\Psi_t\le\psi$ implies
$|g_{t,i}|\le\max\{\sqrt{\hat d_{\max}\psi},\psi/\rho\}$ for every
coordinate $i$.
\end{remark}

\subsection{Local linear convergence}
\label{app:proofs:local-rate}

The global result above does not require clipping to be inactive.
Once the optimization enters a sufficiently local region, however,
the preconditioned gradient becomes small and clipping no longer
activates. In this regime we obtain a sharper linear rate.

\begin{assumption}[Preconditioned Polyak--Lojasiewicz condition]
\label{ass:preconditioned-pl}
There exists $\mu_D\in(0,\alpha]$ such that, whenever $\theta_t$ lies in
a neighborhood $\mathcal N$ of the target solution,
\begin{equation}
    g_t^\top
    D_t^{-1}
    g_t
    \ge
    2\mu_D
    \left(
        F(\theta_t)-F_\star
    \right).
    \label{eq:preconditioned-pl}
\end{equation}
\end{assumption}

\begin{restated}{Theorem}{thm:main-linear}[Local linear convergence with approximate curvature]
Suppose Assumptions
\ref{ass:preconditioned-smoothness} and
\ref{ass:preconditioned-pl} hold, and
\begin{equation}
    (1-\delta)D_t
    \preceq
    \widehat D_t
    \preceq
    (1+\delta)D_t,
    \qquad
    0\le\delta<1.
\end{equation}
Suppose further that
\begin{equation}
    \|
        D_t^{-1}g_t
    \|_\infty
    \le
    (1-\delta)\rho.
    \label{eq:no-clipping-condition}
\end{equation}
Then clipping is inactive and
\begin{equation}
    s_t
    =
    \widehat D_t^{-1}g_t.
\end{equation}
For any
\begin{equation}
    0<\eta<
    \frac{
        2(1-\delta)
    }{\alpha},
\end{equation}
we have
\begin{align}
    F(\theta_{t+1})-F_\star
    &\le
    \left[
        1
        -
        \frac{
            2\eta\mu_D
        }{
            1+\delta
        }
        \left(
            1-
            \frac{
                \alpha\eta
            }{
                2(1-\delta)
            }
        \right)
    \right]
    \left(
        F(\theta_t)-F_\star
    \right).
    \label{eq:general-local-contraction}
\end{align}
In particular, choosing
\begin{equation}
    \eta
    =
    \frac{
        1-\delta
    }{\alpha}
    \label{eq:chosen-local-lr}
\end{equation}
gives
\begin{equation}
    F(\theta_{t+1})-F_\star
    \le
    \left[
        1
        -
        \frac{\mu_D}{\alpha}
        \frac{
            1-\delta
        }{
            1+\delta
        }
    \right]
    \left(
        F(\theta_t)-F_\star
    \right).
    \label{eq:local-contraction}
\end{equation}
\end{restated}

\emph{Proof.}
From
$\widehat D_t\succeq(1-\delta)D_t$,
\begin{equation}
    \frac{|g_{t,i}|}{\hat d_{t,i}}
    \le
    \frac{
        |g_{t,i}|
    }{
        (1-\delta)d_{t,i}
    }.
\end{equation}
Condition~\eqref{eq:no-clipping-condition} therefore implies
\begin{equation}
    \frac{|g_{t,i}|}{\hat d_{t,i}}
    \le\rho,
\end{equation}
so clipping is inactive.

Using Theorem~\ref{thm:main-descent} with
$s_t=\widehat D_t^{-1}g_t$,
\begin{align}
    F(\theta_{t+1})
    &\le
    F(\theta_t)
    -
    \eta
    \left(
        1-
        \frac{
            \alpha\eta
        }{
            2(1-\delta)
        }
    \right)
    g_t^\top
    \widehat D_t^{-1}
    g_t.
    \label{eq:local-step1}
\end{align}
Since
\begin{equation}
    \widehat D_t
    \preceq
    (1+\delta)D_t,
\end{equation}
inversion reverses the PSD order:
\begin{equation}
    \widehat D_t^{-1}
    \succeq
    \frac{
        1
    }{
        1+\delta
    }
    D_t^{-1}.
\end{equation}
Hence
\begin{align}
    g_t^\top
    \widehat D_t^{-1}
    g_t
    &\ge
    \frac{
        1
    }{
        1+\delta
    }
    g_t^\top
    D_t^{-1}
    g_t\\
    &\ge
    \frac{
        2\mu_D
    }{
        1+\delta
    }
    \left(
        F(\theta_t)-F_\star
    \right),
    \label{eq:local-step2}
\end{align}
where the second inequality uses
Assumption~\ref{ass:preconditioned-pl}.
Substituting
\eqref{eq:local-step2}
into
\eqref{eq:local-step1}
proves
\eqref{eq:general-local-contraction}.
Setting
$\eta=(1-\delta)/\alpha$
gives
\eqref{eq:local-contraction}.
\hfill$\square$

Theorem~\ref{thm:main-linear} makes the effect of curvature estimation
error explicit. For exact reference curvature, $\delta=0$, the
contraction coefficient is
\begin{equation}
    1-\frac{\mu_D}{\alpha}.
\end{equation}
With approximate ZO curvature, the effective progress coefficient is
reduced continuously by
\begin{equation}
    \frac{1-\delta}{1+\delta}.
    \label{eq:estimation-penalty}
\end{equation}
Thus curvature estimation error shrinks the stable stepsize and slows
the contraction rate, but does not destroy convergence as long as
$\delta<1$.

\subsection{Convergence of ZO-Curv}
\label{app:proofs:query-rate}

We finally combine the estimator and optimization results over a horizon
of $T$ iterations, with one curvature estimate per training step.
Throughout, $\Lambda_T$, $R_T$, $\vartheta_T$, $\varrho_T$, $t_0$, and
$\delta_T$ are as in Lemma~\ref{lem:ema-error}.

\begin{theorem}[Convergence of ZO-Curv]
\label{thm:query-dependent}
Suppose Assumptions
\ref{ass:hessian-lipschitz} (with $R=R_T$),
\ref{ass:slow-curvature}, and
\ref{ass:preconditioned-smoothness}
hold over a horizon of $T$ iterations, and $q\ge\Lambda_T$.
Then, with probability at least $1-\zeta$, simultaneously for every
iteration $t_0\le t<T$,
\begin{equation}
    (1-\delta_T)D_t
    \preceq
    \widehat D_t
    \preceq
    (1+\delta_T)D_t.
    \label{eq:uniform-distortion}
\end{equation}
If
\begin{equation}
    \delta_T<1
\end{equation}
and
\begin{equation}
    \eta
    \le
    \frac{
        1-\delta_T
    }{\alpha},
\end{equation}
then
\begin{equation}
    \min_{t_0\le t<T}\Psi_t
    \le
    \frac{
        2(
            F(\theta_{t_0})-F_\star
        )
    }{
        \eta (T-t_0)
    }.
    \label{eq:query-stationarity}
\end{equation}

If, in addition, the iterates $\theta_{t_0},\ldots,\theta_{T-1}$ remain in
the neighborhood $\mathcal N$ of Assumption~\ref{ass:preconditioned-pl}
and satisfy the no-clipping condition~\eqref{eq:no-clipping-condition}
with $\delta$ replaced by $\delta_T$, then choosing
\begin{equation}
    \eta
    =
    \frac{
        1-\delta_T
    }{\alpha}
\end{equation}
gives
\begin{equation}
    F(\theta_T)-F_\star
    \le
    \left[
        1
        -
        \frac{\mu_D}{\alpha}
        \frac{
            1-\delta_T
        }{
            1+\delta_T
        }
    \right]^{T-t_0}
    \left(
        F(\theta_{t_0})-F_\star
    \right).
    \label{eq:query-linear-rate}
\end{equation}
\end{theorem}

\emph{Proof.}
By Lemma~\ref{lem:ema-error}, with probability at least $1-\zeta$,
$(1-\delta_T)\tilde v_t\le v_t^{\mathrm{ZO}}\le(1+\delta_T)\tilde v_t$
coordinate-wise for every $t_0\le t\le T$.
Lemma~\ref{lem:denominator-distortion} with $\delta=\delta_T$ then
proves~\eqref{eq:uniform-distortion}.
Equation~\eqref{eq:query-stationarity} follows from
Corollary~\ref{cor:stationarity} applied to the iterations
$t_0,\ldots,T-1$, and~\eqref{eq:query-linear-rate} follows from
Theorem~\ref{thm:main-linear} applied at each of these iterations.
\hfill$\square$

The query budget enters Theorem~\ref{thm:query-dependent} through the
reference preconditioner rather than through $\delta_T$: $D_t$ is built
from the damped target $\tilde h_t$, whose damping
$\lambda_t=\|h_t^{\mathrm{G}}\|_1/q$ decreases as $1/q$, so larger
budgets bring the reference update closer to the undamped
GNB-preconditioned update of Sophia-G. The estimation error $\delta_T$
is controlled by the EMA window and depends on $d$, $q$, and $T$ only
logarithmically.

\begin{corollary}[Controlled penalty relative to reference GNB curvature]
\label{cor:oracle-recovery}
Suppose there exists a constant $\bar\delta<1$ such that
$\delta_T\le\bar\delta$.
Then the number of local iterations $T-t_0$ required to reach
\begin{equation}
    F(\theta_T)-F_\star
    \le
    \epsilon_{\mathrm{opt}}
\end{equation}
satisfies
\begin{equation}
    T-t_0
    =
    O\left(
        \frac{\alpha}{\mu_D}
        \frac{
            1+\bar\delta
        }{
            1-\bar\delta
        }
        \log
        \frac{
            F(\theta_{t_0})-F_\star
        }{
            \epsilon_{\mathrm{opt}}
        }
    \right).
    \label{eq:iteration-complexity}
\end{equation}
For exact reference curvature, $\bar\delta=0$, this reduces to
\begin{equation}
    T_{\mathrm{G}}
    =
    O\left(
        \frac{\alpha}{\mu_D}
        \log
        \frac{
            F(\theta_{t_0})-F_\star
        }{
            \epsilon_{\mathrm{opt}}
        }
    \right).
\end{equation}
Hence the multiplicative iteration penalty of ZO curvature relative
to the reference GNB-preconditioned core is bounded by
\begin{equation}
    \frac{
        T_{\mathrm{ZO}}
    }{
        T_{\mathrm{G}}
    }
    \lesssim
    \frac{
        1+\bar\delta
    }{
        1-\bar\delta
    }.
    \label{eq:relative-iteration-penalty}
\end{equation}
For small $\bar\delta$,
\begin{equation}
    \frac{
        1+\bar\delta
    }{
        1-\bar\delta
    }
    =
    1+2\bar\delta
    +
    O(\bar\delta^2).
    \label{eq:small-error-expansion}
\end{equation}
\end{corollary}

\emph{Proof.}
From Theorem~\ref{thm:query-dependent},
\begin{equation}
    F(\theta_T)-F_\star
    \le
    (1-\nu)^{T-t_0}
    \left(
        F(\theta_{t_0})-F_\star
    \right),
\end{equation}
where
\begin{equation}
    \nu
    :=
    \frac{\mu_D}{\alpha}
    \frac{
        1-\bar\delta
    }{
        1+\bar\delta
    }.
\end{equation}
Using
\begin{equation}
    1-\nu
    \le
    e^{-\nu},
\end{equation}
it is sufficient that
\begin{equation}
    T-t_0
    \ge
    \frac{1}{\nu}
    \log
    \frac{
        F(\theta_{t_0})-F_\star
    }{
        \epsilon_{\mathrm{opt}}
    },
\end{equation}
which gives
\eqref{eq:iteration-complexity}.
Equation~\eqref{eq:small-error-expansion}
is the Taylor expansion around $\bar\delta=0$.
\hfill$\square$

\begin{remark}[Pipelined updates with staleness $s$]
\label{rem:staleness}
In the deployed pipeline (Section~\ref{sec:setting} and
Algorithm~\ref{alg:zocurv}), the EMA $v_t^{\mathrm{ZO}}$ used at step $t$
is built from estimates $h_{t'}^{\mathrm{ZO}}$ queried at earlier
snapshots $\theta_{t'}$, $t'\le t-s$, and is held fixed between
refreshes. Define the reference EMA $\tilde v_t$ with exactly the same
schedule, from $\tilde h_{t'}$ at the same snapshots, mini-batches, and
sampled labels. Proposition~\ref{prop:main-curvature} is a per-refresh
statement. Lemma~\ref{lem:ema-error} applies to the sequence of
refreshes, with $s$ indexing refreshes and $\mathcal G_s$ the history
before the directions of refresh $s$: these directions are still drawn
independently of $\mathcal G_s$, and a held step copies both EMA states.
Lemma~\ref{lem:denominator-distortion} depends only on
$v_t^{\mathrm{ZO}}$ and $\tilde v_t$. Theorems~\ref{thm:main-descent}
and~\ref{thm:main-linear} are per-step statements about $D_t$ and
$\widehat D_t$, with Assumptions~\ref{ass:preconditioned-smoothness}
and~\ref{ass:preconditioned-pl} stated relative to the lagged $D_t$.
Theorem~\ref{thm:query-dependent} unions over at most $T$ refreshes.
All results therefore hold verbatim for any staleness $s$, including the
one-step offset of the main experiments and the $s=3$ deployment of
Appendix~\ref{app:async-deploy}; staleness enters only through the
reference $D_t$, which the assumptions already constrain.
\end{remark}

\subsection{Interpretation}
\label{app:proofs:interpretation}

The above results establish the following chain:
\begin{equation}
    q,\varepsilon
    \quad\Longrightarrow\quad
    \tilde h_t
    \ \text{(unbiased target)}
    \quad\Longrightarrow\quad
    \delta_T
    \ \text{(EMA)}
    \quad\Longrightarrow\quad
    \text{optimization rate}.
\end{equation}

The finite query budget has two separate effects. Its systematic effect
is the uniform damping $\lambda_t=\|h_t^{\mathrm{G}}\|_1/q$ of the
curvature target (Proposition~\ref{prop:main-curvature}), which
averaging before squaring keeps a factor $q$ smaller than squaring
before averaging, and which vanishes as $q\rightarrow\infty$. Its random
effect is a fluctuation of the order of the target in every coordinate,
with constants independent of $d$. The curvature EMA averages this
fluctuation over its effective window (Lemma~\ref{lem:ema-error}), and
the positive floor preserves the resulting relative bound
(Lemma~\ref{lem:denominator-distortion}), which yields the multiplicative
preconditioner distortion $\delta_T$ of~\eqref{eq:uniform-delta}.
As long as $\delta_T<1$, ZO-Curv preserves descent.
In the local unclipped regime, the convergence coefficient is reduced
from the reference value
\begin{equation}
    \frac{\mu_D}{\alpha}
\end{equation}
to
\begin{equation}
    \frac{\mu_D}{\alpha}
    \frac{
        1-\delta_T
    }{
        1+\delta_T
    },
\end{equation}
where $\mu_D$ and $\alpha$ refer to the damped GNB reference
preconditioner.

Therefore, curvature-estimation error produces a continuous and
controlled degradation of the convergence rate rather than destroying
the descent direction. This differs fundamentally from injecting ZO
noise directly into the first-order gradient: an erroneous gradient
estimate can rotate or reverse the descent direction, whereas a
positive curvature estimate only rescales the coordinate-wise
first-order update, with clipping additionally bounding the maximum
step magnitude.

Thus, replacing the backpropagation-based GNB computation with
forward-only ZO queries yields a damped GNB preconditioner: the query
budget controls the damping, the curvature EMA controls the estimation
error, and the convergence mechanism of clipped curvature
preconditioning is retained.

%% file: floats/tab_hparams.tex
\begin{table}[h]
\caption{\textbf{Model configurations and peak learning rates.} \ours{}
uses the SophiaG hyperparameters unchanged; $\gamma$ is the curvature
scale of~\eqref{eq:zo-update}.}
\label{tab:hparams}
\begin{center}
\small
\begin{tabular}{lcccccc}
\toprule
Model & $d_{\mathrm{model}}$ & Heads & Depth & AdamW lr
& SophiaG / \ours{} lr & $\gamma$ \\
\midrule
GPT-2 Small (125M)  & 768  & 12 & 12 & $6\times10^{-4}$ & $6\times10^{-4}$ & 0.05 \\
GPT-2 Medium (355M) & 1024 & 16 & 24 & $3\times10^{-4}$ & $7\times10^{-4}$ & 0.08 \\
GPT-2 Large (770M)  & 1280 & 20 & 36 & $2\times10^{-4}$ & $3\times10^{-4}$ & 0.05 \\
GPT-NeoX (1.5B)     & 1536 & 24 & 48 & $1.5\times10^{-4}$          & $1.2\times10^{-4}$ & 0.05 \\
\bottomrule
\end{tabular}
\end{center}
\end{table}

%% file: floats/fig3_scaling.tex
\begin{figure}[h]
\centering
\includegraphics[width=\textwidth]{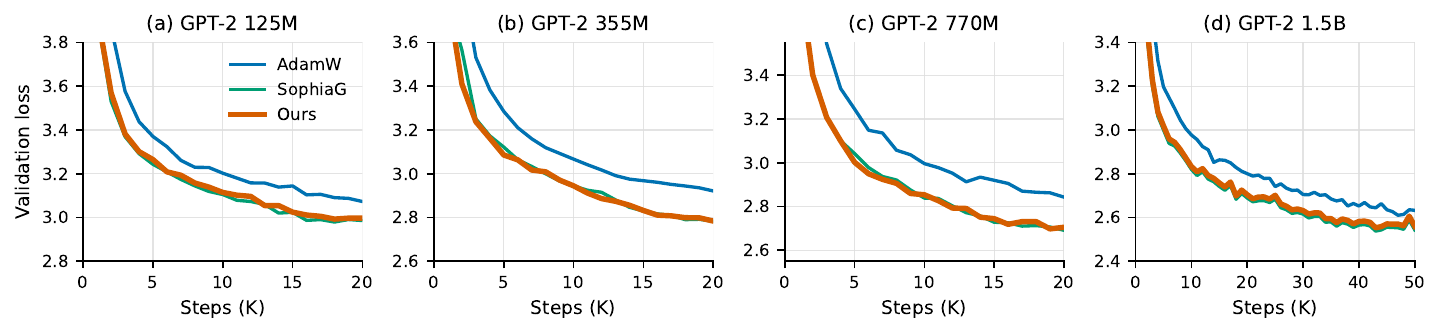}
\caption{\textbf{\ours{} matches SophiaG's step efficiency at every
scale} (first 20K steps of the runs of Figure~\ref{fig:main}). The two
curves are nearly indistinguishable on a step axis: the forward-only
estimator matches the backpropagated one in per-step quality, and the
advantage of \ours{} lies in where the estimate is computed.}
\label{fig:scaling}
\end{figure}

%% file: floats/fig_sophia_k10.tex
\begin{figure}[t]
\centering
\includegraphics[width=\textwidth]{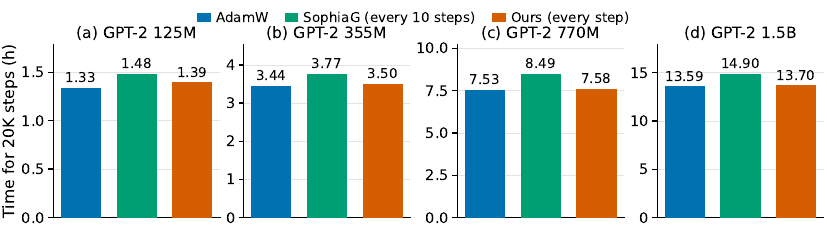}
\caption{\textbf{SophiaG with a ten-step refresh interval.} Wall-clock
time to complete 20K steps for AdamW, SophiaG refreshing its curvature
every ten steps, and \ours{} refreshing at every step. Even with its
curvature cost amortized, SophiaG is slower than \ours{} at every size.}
\label{fig:sophia-k10}
\end{figure}

%% file: floats/fig_sophia_k10_loss.tex
\begin{figure}[t]
\centering
\includegraphics[width=\textwidth]{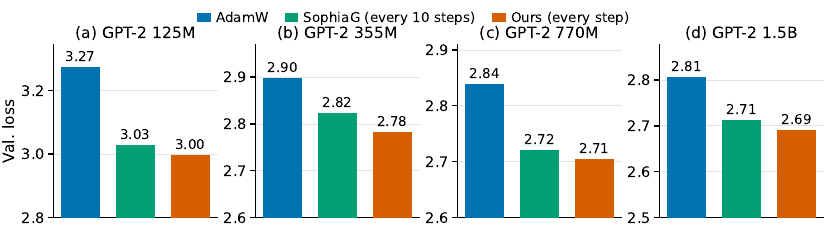}
\caption{\textbf{Loss of SophiaG with a ten-step refresh interval.}
Validation loss after 20K steps for the runs of
Figure~\ref{fig:sophia-k10}. With its curvature refreshed ten times less
often, SophiaG ends above \ours{} at every size.}
\label{fig:sophia-k10-loss}
\end{figure}

%% file: floats/fig7_query_curves.tex
\begin{figure}[h]
\centering
\includegraphics[width=\textwidth]{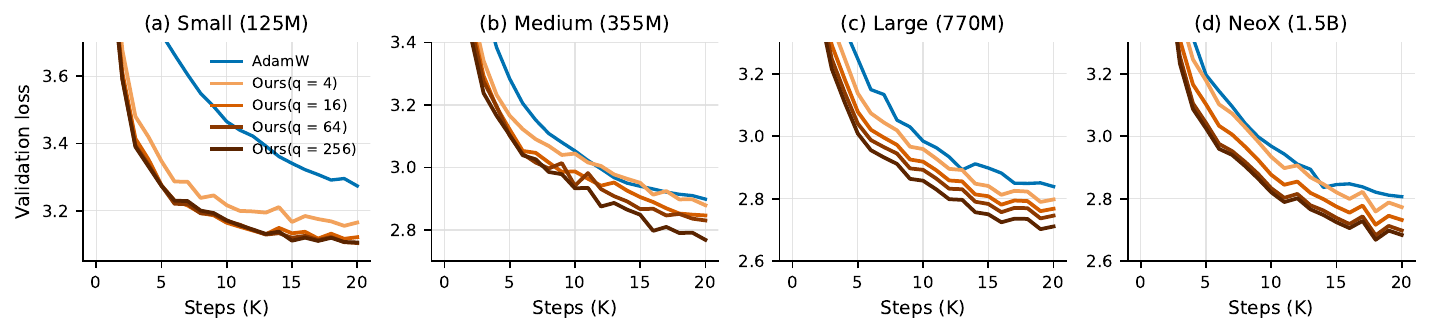}
\caption{\textbf{Training trajectories across query budgets, all four
sizes} (companion to Figure~\ref{fig:query}). At 125M the budgets
collapse onto each other early, whereas at 770M and 1.5B they remain
cleanly ordered, matching the scale-dependent saturation of
Figure~\ref{fig:query}(a,b).}
\label{fig:query-curves}
\end{figure}

%% file: floats/fig9_pipeline_offset.tex
\begin{figure}[h]
\centering
\includegraphics[width=0.55\textwidth]{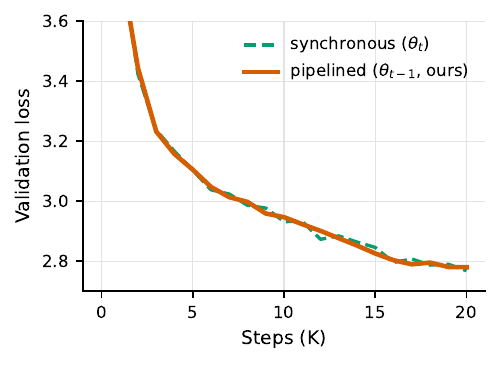}
\caption{\textbf{The one-step pipeline offset is free} (GPT-2 355M,
$\zoq=512$). Training with curvature estimated at $\theta_{t-1}$
(pipelined, as deployed) is indistinguishable from using curvature
estimated at the current $\theta_t$ (synchronous).}
\label{fig:pipeline-offset}
\end{figure}

%% file: floats/tab2_downstream.tex
\begin{table}[h]
\caption{\textbf{Full downstream evaluation} of the GPT-2 770M checkpoints
at 20K steps (companion to Figure~\ref{fig:downstream}). Accuracy in \%.
Base: few-shot before finetuning (10 examples for HellaSwag, 25 for
ARC-E and ARC-C, none for LAMBADA). FT: zero-shot after per-task LoRA
finetuning. LAMBADA has no public training split and is not finetuned;
it enters the FT average with its untuned score.}
\label{tab:downstream}
\begin{center}
\small
\begin{tabular}{lcccccccccc}
\toprule
& \multicolumn{2}{c}{HellaSwag} & \multicolumn{2}{c}{ARC-E}
& \multicolumn{2}{c}{ARC-C} & \multicolumn{2}{c}{LAMBADA}
& \multicolumn{2}{c}{Average} \\
\cmidrule(lr){2-3}\cmidrule(lr){4-5}\cmidrule(lr){6-7}
\cmidrule(lr){8-9}\cmidrule(lr){10-11}
Method & Base & FT & Base & FT & Base & FT & Base & FT
& Base & FT \\
\midrule
AdamW   & 31.4 & 33.0 & 40.2 & 50.9 & 22.1 & 26.1
        & 30.8 & -- & 31.1 & 35.2 \\
SophiaG & 35.6 & 37.1 & 47.5 & 54.7 & 25.8 & 28.1
        & 39.1 & -- & 37.0 & 39.8 \\
\textbf{\ours{} (ours)} & 35.2 & 36.3 & 47.4 & 54.2
        & 26.8 & 28.8 & 39.0 & -- & 37.1 & 39.6 \\
\bottomrule
\end{tabular}
\end{center}
\end{table}